\documentclass{article}

\usepackage{arxiv}

\usepackage[utf8]{inputenc} 
\usepackage[T1]{fontenc}    
\usepackage{hyperref}       
\usepackage{url}            
\usepackage{booktabs}       
\usepackage{amsfonts}       
\usepackage{nicefrac}       
\usepackage{microtype}      
\usepackage{lipsum}
\usepackage{graphicx}
\usepackage{color}
\usepackage{amssymb}

\usepackage{amsmath,amsfonts}
\usepackage{algorithmic}
\usepackage{algorithm}
\usepackage{array}
\usepackage{subfig}
\usepackage{textcomp}
\usepackage{stfloats}
\usepackage{url}
\usepackage{verbatim}
\usepackage{graphicx}
\usepackage{cite}
\usepackage{bm}

\title{Truncated Rectified Flow Policy for Reinforcement Learning with One-Step Sampling}

\author{
 Xubin Zhou \\
  Harbin Institute of Technology\\
  \texttt{24B904016@stu.hit.edu.cn} \\
   \And
 Yipeng Yang \\
  Harbin Institute of Technology\\
  \texttt{ypyang@hit.edu.cn} \\
  \And
 Zhan Li \\
  Harbin Institute of Technology\\
  \texttt{zhanli@hit.edu.cn} \\
}

\begin{document}
\maketitle
\begin{abstract}
Maximum entropy reinforcement learning (MaxEnt RL) has become a standard framework for sequential decision making, yet its standard Gaussian policy parameterization is inherently unimodal, limiting its ability to model complex multimodal action distributions.
This limitation has motivated increasing interest in generative policies based on diffusion and flow matching as more expressive alternatives.
However, incorporating such policies into MaxEnt RL is challenging for two main reasons: the likelihood and entropy of continuous-time generative policies are generally intractable, and multi-step sampling introduces both long-horizon backpropagation instability and substantial inference latency.
To address these challenges, we propose Truncated Rectified Flow Policy (TRFP), a framework built on a hybrid deterministic-stochastic architecture.
This design makes entropy-regularized optimization tractable while supporting stable training and effective one-step sampling through gradient truncation and flow straightening.
Empirical results on a toy multigoal environment and 10 MuJoCo benchmarks show that TRFP captures multimodal behavior effectively, outperforms strong baselines on most benchmarks under standard sampling, and remains highly competitive under one-step sampling.
\end{abstract}


\section{Introduction}
\label{sec:introduction}
Reinforcement learning has established itself as a foundational framework for solving complex sequential decision problems \cite{lian2025reinforcement, gao2026amplitude}. Within this framework, maximum entropy reinforcement learning (MaxEnt RL) modifies the traditional optimization goal by integrating the policy entropy into the reward signal. This regularization encourages the agent to maintain diverse action choices, which directly improves adaptability and robustness during training. However, most existing approaches within this framework rely on Gaussian distributions for policy parameterization. The inherent unimodality of Gaussian models severely limits the representational capacity of the policy. The agent struggles to model complex multimodal action distributions, resulting in suboptimal performance across challenging environments \cite{chen2025compact}.

To address these bottlenecks, generative models based on diffusion \cite{ho2020ddpm,song2021sde} and flow matching \cite{lipman2023flowmatching,liu2023rectifiedflow}, have recently emerged as highly effective alternatives for policy parameterization. These generative policies exhibit a superior capacity for modeling complex multimodal distributions, thereby showing strong potential in domains such as autonomous driving\cite{gui2025trajdiff,li2025recogdrive,gui2026worlddrive} and robotics\cite{zhou2026humanoidmamba,chi2023diffusionpolicy,chisari2024pointflowmatch,zhou2026cgcl}.
Despite their remarkable expressiveness, integrating these generative policies into MaxEnt RL faces two fundamental obstacles.
The first major impediment is the theoretical intractability of likelihood computation. Standard MaxEnt algorithms rely on evaluating the exact log-likelihood of the policy to estimate the entropy. However, deriving the exact marginal likelihood for diffusion models or flow matching typically involves solving complex differential equations, rendering the process computationally infeasible during online training.
The second major obstacle relates to the computational burden of optimization and inference. Generative models fundamentally rely on a multi-step iterative process for sampling. The direct application of backpropagation through time (BPTT) across the entirety of this generation chain frequently results in severe vanishing or exploding gradients. Furthermore, this iterative requirement introduces substantial latency during inference, thereby preventing the policy from satisfying the demands of real-time control applications.

To overcome the aforementioned challenges, we propose a novel MaxEnt RL framework termed Truncated Rectified Flow Policy (TRFP).
The proposed framework integrates three fundamental mechanisms to achieve stable training. First, we introduce a hybrid architecture that separates the overall process of sampling into a prefix governed by a deterministic ordinary differential equation (ODE) and a tail governed by a stochastic differential equation (SDE). 
This structural design enables a tractable surrogate log-likelihood to be formulated along the generation path. As a result, it bypasses the computationally expensive integration of the velocity-field divergence that is typically required in continuous-flow models. 
Second, the framework incorporates a gradient truncation mechanism at the prefix-tail interface. By restricting optimization exclusively to the stochastic refinement phase, this approach prevents gradient pathologies associated with BPTT over long horizons.
Unlike DRaFT \cite{clark2024draft}, which applies truncated backpropagation to reward-based finetuning of pre-trained text-to-image diffusion models, our truncation serves as part of a decoupled MaxEnt RL optimization scheme for rectified flow policies trained from scratch.
Finally, we employ flow straightening regularization for decoupled training, utilizing self-distillation to enforce linear transport trajectories within the prefix. This structural constraint simultaneously provides a theoretical motivation by minimizing the surrogate divergence error and yields a practical advantage by allowing one-step sampling during deployment.

Our core contributions are summarized as follows:
\begin{itemize}
\item We propose the truncated rectified flow policy framework, which integrates flow matching into MaxEnt RL. This framework addresses the dual challenges of intractable likelihood computation and unstable backpropagation through time that typically hinder flow policies in online settings.
\item By incorporating flow straightening regularization, we establish a theoretical boundary for the error of the approximate entropy. In practice, this mechanism enables highly competitive one-step inference with high fidelity.
\item Extensive experiments demonstrate that the proposed method achieves state-of-the-art performance on most MuJoCo benchmark tasks, and attains comparable results to existing diffusion policy methods even under one-step sampling.
\end{itemize}

\section{Related Work}
\label{sec:related_work}

\textbf{Maximum Entropy Reinforcement Learning.}
MaxEnt RL is one of the most widely adopted off-policy RL frameworks due to beneficial exploration properties.
Represented by the SAC \cite{haarnoja2018sac}, this approach establishes a robust off-policy actor-critic paradigm. By explicitly integrating policy entropy alongside the cumulative return, it achieves a superior balance among exploration capacity, sample efficiency, and training stability.
Recent advanced baselines including CrossQ \cite{bhatt2024crossq} and BRO \cite{nauman2024bro} further advance sample efficiency and computational performance in continuous control. However, these improvements primarily originate from critic network normalization, structural regularization, and architectural scaling, rather than expanding policy representational capacity.
Motivated by this context, replacing the conventional Gaussian policy with highly expressive generative policy, while preserving the inherent advantages of MaxEnt RL, emerges as a critical direction for future research.

\textbf{Generative Policies in Imitation Learning and Offline RL.}
Recent advances in generative models have made generative policies attractive alternatives to Gaussian parameterizations. In imitation learning and offline RL, these models have shown strong ability to capture complex multimodal action distributions, leading to improved policy expressiveness and strong empirical performance \cite{wang2023diffusionql,chen2023sfbc,fang2025flowpolicy,park2025fql,hansen2023idql}. These results establish generative policies as a promising foundation for decision making, but they do not directly resolve the additional challenges of online optimization, where the policy must be updated from evolving value estimates while maintaining efficient sampling and stable training.

\textbf{Diffusion and Flow Policies in Online Reinforcement Learning.}
In contrast to offline RL and imitation learning, online RL introduces a fundamental challenge for generative policy optimization. Since the agent must improve its policy from rewards collected through environment interaction, rather than relying on directly supervised target actions, diffusion and flow policies can no longer depend on standard behavior cloning objectives.
To address this issue, existing methods mainly focus on translating Q-function information into trainable updates for generative policies without access to supervised action targets.
DIPO \cite{yang2023dipo} updates replay-buffer actions using the action gradient of the Q-function, and then trains a diffusion policy to imitate the improved actions.
QSM \cite{psenka2024qsm} instead links the policy score to the action gradient of the Q-function, casting policy updates through the lens of score matching.
QVPO \cite{ding2024qvpo} and RSM \cite{ma2025rsm} incorporate value information into diffusion training through weighted objectives, respectively via a Q-weighted variational objective and reweighted score matching, thereby avoiding direct supervision from optimal action targets.

A second core difficulty lies in the intractability of likelihood and entropy, which forms a central theoretical bottleneck when combining generative policies with MaxEnt RL.
Consequently, generative policies cannot be directly incorporated into standard MaxEnt and policy-gradient methods, such as SAC and PPO \cite{schulman2017ppo}, which typically require tractable action likelihoods.
DACER \cite{wang2024dacer} addresses this issue by estimating the entropy of the diffusion policy with a Gaussian mixture surrogate.
DIME \cite{celik2025dime} approaches this challenge from an approximate inference perspective by deriving a lower bound on the MaxEnt objective.
MaxEntDP \cite{dong2025maxentdp} approximates the log-probability of the diffusion policy within the SAC framework through Q-weighted noise estimation and numerical integration.
Beyond off-policy methods, recent work has also investigated how diffusion policies can be optimized in on-policy settings. 
DPPO \cite{ren2025dppo} treats the denoising chain as a diffusion MDP with Gaussian transitions, while ReinFlow \cite{zhang2025reinflow} injects learnable noise into deterministic flow trajectories to convert them into a discrete-time Markov process.
GenPO \cite{ding2025genpo} further leverages exact diffusion inversion to construct an invertible action mapping, thereby enabling action likelihood estimation.

Beyond the challenges of objective design and likelihood tractability, iterative generative policies face an additional critical bottleneck. Specifically, the multi-step process of denoising or flow rollout introduces substantial inference latency, and it frequently causes optimization instability when gradients are backpropagated through the complete generation chain.
In fine-tuning of text-to-image diffusion models, DRaFT \cite{clark2024draft} improves the efficiency of long-chain optimization through truncated backpropagation, together with LoRA and gradient checkpointing.
SAC Flow \cite{zhang2025sacflow} interprets multi-step flow rollout as a form of residual recurrent computation, attributing its training instability to the resulting vanishing and exploding gradients.
FPMD \cite{chen2025fpmd} moves toward one-step inference for flow policies, thereby reducing the sampling overhead of generative policies in online decision making.
Overall, these studies suggest that the continued development of online generative policies depends not only on tractable likelihood estimation, but also on achieving efficient sampling and stable optimization over multi-step generation processes.

\section{Preliminaries}
\label{sec:preliminaries}

\subsection{Maximum Entropy Reinforcement Learning}
We formulate the sequential decision-making problem as an infinite-horizon Markov Decision Process (MDP), defined by the tuple $(\mathcal{S}, \mathcal{A}, p, r, \gamma)$, where $\mathcal{S} \in \mathbb{R}^{d_s}$ denotes the state space, and $\mathcal{A} \in \mathbb{R}^{d_a}$ represents the continuous action space. At each discrete time step $t$, the agent selects an action $\mathbf{a}_t$ based on the current state $\mathbf{s}_t$ according to a policy $\pi(\cdot|\mathbf{s}_t)$. The environment then transitions to the next state $\mathbf{s}_{t+1}$ governed by the dynamics $p(\mathbf{s}_{t+1}|\mathbf{s}_t, \mathbf{a}_t)$ and yields a scalar reward $r(\mathbf{s}_t, \mathbf{a}_t)$.

Unlike standard reinforcement learning, which maximizes the expected cumulative reward, MaxEnt RL augments the objective with an entropy term $H(\pi(\cdot|\mathbf{s}_t)) = \mathbb{E}_{\mathbf{a} \sim \pi}[-\log \pi(\mathbf{a}|\mathbf{s}_t)]$. The primary goal is to learn an optimal policy $\pi^*$ that maximizes the expected entropy-regularized return:
\begin{equation}
    J(\pi) = \mathbb{E}_{\tau \sim \pi} \left[ \sum_{t=0}^{\infty} \gamma^t \left( r(\mathbf{s}_t, \mathbf{a}_t) + \alpha H(\pi(\cdot|\mathbf{s}_t)) \right) \right],
    \label{eq:maxent_objective}
\end{equation}
where $\gamma \in [0, 1)$ is the discount factor, and $\alpha > 0$ is the temperature parameter that balances the trade-off between exploration and exploitation.

SAC \cite{haarnoja2018sac} is a prominent off-policy algorithm designed to optimize this objective via an iterative process. In the policy evaluation step, SAC learns a parameterized soft Q-function $Q_{\psi}(\mathbf{s}, \mathbf{a})$ by minimizing the soft Bellman residual:
\begin{equation}
    \mathcal{L}_{Q}(\psi) = \mathbb{E}_{(\mathbf{s}, \mathbf{a}, r, \mathbf{s}') \sim \mathcal{D}} \left[ \left( Q_{\psi}(\mathbf{s}, \mathbf{a}) - y \right)^2 \right],
    \label{eq:critic_loss}
\end{equation}
where $y = r + \gamma \mathbb{E}_{\mathbf{a}' \sim \pi_{\theta}(\cdot|\mathbf{s}')} \left[ \min_{j=1,2} Q_{\bar{\psi}_j}(\mathbf{s}', \mathbf{a}') - \alpha \log \pi_{\theta}(\mathbf{a}'|\mathbf{s}') \right]$ is the soft target value constructed using target networks $\bar{\psi}$.

In the policy improvement step, the policy parameters $\theta$ are updated to minimize the Kullback-Leibler divergence between the policy and the Boltzmann distribution induced by the Q-function:
\begin{equation}
    \mathcal{L}_{\pi}(\theta) = \mathbb{E}_{\mathbf{s} \sim \mathcal{D}} \left[ \mathbb{E}_{\mathbf{a} \sim \pi_\theta(\cdot|\mathbf{s})} \left[ \alpha \log \pi_\theta(\mathbf{a}|\mathbf{s}) - Q_{\psi}(\mathbf{s}, \mathbf{a}) \right] \right].
    \label{eq:sac_loss}
\end{equation}

Finally, to avoid manual tuning of the temperature parameter, SAC formulates entropy maximization as a constrained optimization problem, where the average entropy is constrained to be at least a target value $\bar{\mathcal{H}}$. This leads to the following dual objective for learning $\alpha$:
\begin{equation}
    \mathcal{L}_{\alpha}(\alpha) = \mathbb{E}_{\mathbf{s} \sim \mathcal{D}, \mathbf{a} \sim \pi_\theta(\cdot|\mathbf{s})} \left[ -\alpha \left( \log \pi_\theta(\mathbf{a}|\mathbf{s}) + \bar{\mathcal{H}} \right) \right].
    \label{eq:alpha_loss}
\end{equation}

While Gaussian policies are standard in SAC, their inherent unimodality limits the agent's ability to capture complex, multimodal action distributions. Our work addresses this limitation by integrating flow-based generative models within the MaxEnt framework, thereby enhancing expressivity while maintaining stable training.

\subsection{From Diffusion to Flow Matching}
\label{sec:flow_matching}

Generative modeling in continuous control tasks aims to approximate the policy distribution $\pi(\mathbf{a}|\mathbf{s})$ using a parameterized model. Diffusion Probabilistic Models (DPMs) \cite{ho2020ddpm,song2021sde} have emerged as a powerful paradigm for this purpose. Typically, DPMs define a forward SDE that gradually corrupts data into Gaussian noise. Sampling is then performed by solving the reverse-time SDE or its equivalent Probability Flow ODE (PF-ODE):
\begin{equation}
    \mathrm{d}\mathbf{x} = \left[ \mathbf{f}(\mathbf{x}, \tau) - \frac{1}{2}g^2(\tau)\nabla_\mathbf{x} \log p_\tau(\mathbf{x}) \right] \mathrm{d}\tau,
\end{equation}
where $\mathbf{f}(\cdot)$ and $g(\cdot)$ denote the drift and diffusion coefficients, respectively. While effective, standard diffusion paths (e.g., Variance Preserving) often induce high-curvature trajectories, necessitating computationally expensive numerical solvers with many steps for high-fidelity sampling.

To overcome the sampling inefficiency of diffusion models, Flow Matching (FM) \cite{lipman2023flowmatching} introduces a generalized framework for training Continuous Normalizing Flows (CNFs) without explicitly simulating the forward SDE. FM directly regresses a time-dependent vector field $\mathbf{v}_\theta(\mathbf{x}, \tau)$ that generates a probability density path $p_\tau(\mathbf{x})$ interpolating between a simple prior $p_0 = \mathcal{N}(\mathbf{0}, \mathbf{I})$ and the data distribution $p_1$. The dynamics are governed by the ODE:
\begin{equation}
    \frac{\mathrm{d}\mathbf{x}}{\mathrm{d}\tau} = \mathbf{v}_\theta(\mathbf{x}_\tau, \tau), \quad \tau \in [0, 1].
    \label{eq:fm_ode}
\end{equation}
The core objective of FM is to match the model vector field $\mathbf{v}_\theta$ to a target conditional vector field $\mathbf{u}_\tau(\mathbf{x}|\mathbf{x}_1)$ that generates the conditional probability path $p_\tau(\mathbf{x}|\mathbf{x}_1)$:
\begin{equation}
    \mathcal{L}_{\text{CFM}}(\theta) = \mathbb{E}_{\tau, \mathbf{x}_1, p_\tau(\mathbf{x}|\mathbf{x}_1)} \left[ \| \mathbf{v}_\theta(\mathbf{x}, \tau) - \mathbf{u}_\tau(\mathbf{x}|\mathbf{x}_1) \|^2 \right].
\end{equation}

While FM allows for arbitrary probability paths, the choice of path significantly impacts sampling efficiency. In this work, we adopt the specific parameterization used in Optimal Transport CFM (OT-CFM) and Rectified Flow \cite{liu2023rectifiedflow}, which constructs the path via linear interpolation:
\begin{equation}
    \mathbf{x}_\tau = \tau \mathbf{x}_1 + (1 - \tau) \mathbf{x}_0, \quad \mathbf{x}_0 \sim p_0, \mathbf{x}_1 \sim p_{data}.
    \label{eq:linear_interp}
\end{equation}
Under this linear design, the conditional target vector field admits a remarkably simple constant form: $\mathbf{u}_\tau(\mathbf{x}|\mathbf{x}_1, \mathbf{x}_0) = \mathbf{x}_1 - \mathbf{x}_0$. This formulation corresponds to the optimal transport plan under Euclidean cost, theoretically yielding straight-line trajectories for individual samples. Although the marginal flow field may still exhibit crossings due to multimodal data couplings, the linear interpolation prior serves as a strong inductive bias for straightness. This property is crucial for our method, as it justifies the use of low-order solvers (and potentially one-step generation) during online interactions, addressing the latency bottleneck inherent in curved diffusion paths.

\section{Method}
\label{sec:method}

We propose TRFP, an off-policy actor-critic framework that trains a flow policy from scratch for decision making. TRFP explicitly addresses three fundamental challenges that hinder the integration of diffusion-based policies with MaxEnt RL:
(i) multi-step sampling incurs the high computational cost and gradient instability of BPTT;
(ii) exact policy likelihood for an ODE-based flow is generally intractable without computing Jacobian
determinants or divergence integrals; and
(iii) iterative generation incurs substantial inference latency during decision making.

To overcome these challenges, our methodology integrates a hybrid ODE-SDE sampling architecture with a path-based entropy surrogate and a decoupled truncated gradient estimator.
Furthermore, to reconcile the truncated training with high-fidelity inference, we introduce a flow straightening regularization mechanism. This auxiliary objective aligns the flow with linear trajectory priors, ensuring that the policy maintains superior expressivity and performance even under the stringent constraint of one-step sampling.

\subsection{The Hybrid Policy Architecture}
\label{sec:hybrid_policy}

Deploying diffusion-based policies in off-policy reinforcement learning faces a fundamental dilemma: while multi-step iterative refinement is essential for modeling complex multimodal distributions, it incurs high inference latency and generates deep computational graphs that are hostile to stable backpropagation. To resolve this, we propose a hybrid sampling architecture that decomposes the generation trajectory of total length $K$ into two distinct phases: a deterministic transport phase (first $K-L$ steps) and a stochastic refinement phase (last $L$ steps).

Formally, let $\mathbf{s} \in \mathcal{S}$ denote the state. We discretize the time horizon into $K$ steps $t_k = k/K$ to generate the latent variable $\mathbf{u}_K$. The generation process starts from a Gaussian prior $\mathbf{u}_0 \sim \mathcal{N}(\mathbf{0}, \mathbf{I})$ and evolves as follows:

\textbf{Deterministic Prefix (ODE Transport).} The first $K-L$ steps constitute the transport phase. We model this phase as a deterministic ODE: $\mathrm{d}\mathbf{u} = \mathbf{v}_\theta^{\mathrm{ode}}(\mathbf{s}, \mathbf{u}, \tau)\mathrm{d}\tau$. For the discrete steps $k < K-L$, we perform numerical integration using a second-order solver. Benefiting from the linear trajectory priors induced by Rectified Flow, this deterministic prefix efficiently transports the probability mass from the uninformative prior to the high-density regions of the target action distribution. This property enables the phase to traverse the trajectory with fewer steps (potentially reducing to a single step), thereby significantly lowering the computational overhead compared to stochastic diffusion processes.

\textbf{Stochastic Tail (SDE Refinement).} The final $L$ steps serve as the refinement phase. To inject the necessary stochasticity, we switch the dynamics to a discretized SDE:
\begin{equation}
    \mathbf{u}_{k+1} = \mathbf{u}_k + \mathbf{v}_\theta^{\mathrm{sde}}(\mathbf{s}, \mathbf{u}_k, t_k) \Delta t + \boldsymbol{\sigma}_\theta(\mathbf{s}, \mathbf{u}_k, t_k) \odot \boldsymbol{\varepsilon}_k, \quad \boldsymbol{\varepsilon}_k \sim \mathcal{N}(\mathbf{0}, \mathbf{I}),
    \label{eq:hybrid_transition}
\end{equation}
where $k \in \{K-L, \dots, K-1\}$, $\Delta t = 1/K$, and $\boldsymbol{\sigma}_\theta$ denotes the noise scale parameterized by the policy network. This hybrid design ensures that the policy retains the multimodal generation capability of flow models via the prior $\mathbf{u}_0$, while the stochastic tail induces a tractable transition density, which is essential for computing the entropy surrogate and policy gradients in the subsequent steps.
After the final update step, the terminal variable $u_K$ is taken as the action output of the policy, i.e., $a = u_K$.

\subsection{Approximate Entropy and Log-Likelihood}
\label{sec:approx_entropy}

Standard MaxEnt RL algorithms, such as SAC, necessitate the computation of the log-likelihood $\log \pi(\mathbf{a}|\mathbf{s})$. However, for ODE-based generative policies, calculating the exact marginal log-likelihood requires evaluating the trace of the vector field's Jacobian. 
Exact likelihood evaluation requires integrating the divergence of the vector field along the ODE trajectory. Both exact Jacobian-trace computation and stochastic trace estimation introduce substantial computational overhead, making online training impractical.

To address this challenge, we construct a tractable surrogate log-likelihood by formulating the entropy maximization objective over the augmented latent trajectory space---the joint space of all stochastic variables involved in the generation process. Specifically, based on the hybrid architecture defined in Sec. \ref{sec:hybrid_policy}, we define the surrogate log-density $\log \tilde{\pi}$ as the joint log-probability of the initial prior and the stochastic tail transitions:
\begin{equation}
    \log \tilde{\pi}(\mathbf{u}_{0:K}|\mathbf{s}) \triangleq \log \mathcal{N}(\mathbf{u}_0; \mathbf{0}, \mathbf{I}) + \sum_{k=K-L}^{K} \log \mathcal{N}(\mathbf{u}_{k+1}; \mathbf{u}_k + \mathbf{v}_\theta^{\mathrm{sde}}\Delta t, \boldsymbol{\sigma}_\theta^2 \mathbf{I}).
    \label{eq:surrogate_logp}
\end{equation}
This factorization circumvents the expensive integration over the deterministic ODE prefix. Effectively, this surrogate objective approximates the marginal density by omitting the volume change induced by the deterministic transport.

To validate the rationality of this approximation in TRFP, we quantitatively examine the introduced error. First, we invoke the standard result from Neural ODEs \cite{chen2018neuralode} describing density evolution in continuous flows:

\textbf{Theorem 1 (Instantaneous Change of Variables).} \textit{For a continuous transformation following the ODE $\frac{\mathrm{d}\mathbf{u}(t)}{\mathrm{d}t} = \mathbf{v}(\mathbf{u}(t), t)$, the rate of change of the log-density along the trajectory is equal to the negative divergence of the velocity field:}
\begin{equation}
    \frac{\mathrm{d}}{\mathrm{d}t} \log p_t(\mathbf{u}(t)) = - \mathrm{tr}\left( \frac{\partial \mathbf{v}}{\partial \mathbf{u}}(\mathbf{u}(t), t) \right) = - \nabla_{\mathbf{u}} \cdot \mathbf{v}(\mathbf{u}(t), t).
\end{equation}

By integrating Theorem 1 over the prefix duration, we derive the upper bound of the error introduced by ignoring the prefix volume change:

\textbf{Proposition 1 (Approximation Error Bound).} \textit{Let $\tau = (K-L)/K$ be the end time of the prefix. The approximation error $\Delta_{\mathrm{pre}} \triangleq \log p(\mathbf{u}_\tau|\mathbf{s}) - \log p(\mathbf{u}_0)$ satisfies:}
\begin{equation}
    |\Delta_{\mathrm{pre}}| = \left| \int_0^\tau \nabla_{\mathbf{u}} \cdot \mathbf{v}^{\mathrm{ode}}_\theta(\mathbf{s}, \mathbf{u}(t), t) \mathrm{d}t \right| \le M \tau,
\end{equation}
\textit{provided that $|\nabla_{\mathbf{u}} \cdot \mathbf{v}^{\mathrm{ode}}_\theta| \le M$ along the trajectory.}

Proposition 1 reveals that the approximation error is controlled by the divergence magnitude of the prefix velocity field. Motivated by this observation, we incorporate flow straightening regularization (detailed in Sec. \ref{sec:fm_regularizer}) to encourage nearly linear transport trajectories in the deterministic prefix. 
In the ideal case where the learned prefix velocity becomes approximately independent of $u$ along the transport path, the induced divergence can be substantially reduced, which in turn helps control the surrogate-likelihood approximation error in practice.
Thus, the flow straightening mechanism empirically supports few-step sampling.

Finally, we incorporate this surrogate into the SAC framework. The actor loss $\mathcal{L}_\pi$ is defined as:
\begin{equation}
    \mathcal{L}_\pi(\theta) = \mathbb{E}_{\mathbf{s}, \mathbf{u}_0, \boldsymbol{\varepsilon}} \left[ \alpha \log \tilde{\pi}(\mathbf{u}_{0:K}|\mathbf{s}) - Q_{\psi_i}(\mathbf{s}, \mathbf{a}) \right].
    \label{eq:sacloss}
\end{equation}
This formulation allows TRFP to efficiently update the policy under a surrogate MaxEnt objective.

\subsection{Truncated Gradient Optimization}
\label{sec:truncated_grad}

While the actor objective $\mathcal{L}_\pi$ (Eq. \ref{eq:sacloss}) derived in the previous section is theoretically principled, optimizing it directly via reparameterization involves differentiating through the entire generation chain. For a $K$-step sampler, this is algebraically equivalent to BPTT in a Recurrent Neural Network (RNN) of depth $K$. As $K$ increases, this full-chain differentiation not only limits the batch size due to memory overhead but also induces severe gradient pathologies, destabilizing the training process.

To mitigate these issues, we propose a Truncated Gradient Estimator. Specifically, we detach the computational graph at the interface between the deterministic prefix and the stochastic tail. Let $k_c = K-L$ denote the cutoff step. 
During training, the prefix produces an intermediate state $\mathbf{u}_{k_c}$, which is used to initialize the stochastic tail without allowing gradients to propagate back into the prefix. 
The policy is then optimized only through the SDE dynamics from $\mathbf{u}_{k_c}$ to $\mathbf{u}_K$, such that the SAC objective affects only the final $L$ steps. This truncated optimization avoids long-horizon backpropagation through the deterministic prefix.

This truncation strategy is conceptually related to DRaFT \cite{clark2024draft}, which finetunes pretrained text-to-image diffusion models under differentiable rewards. However, the role of truncation in TRFP is different: it is not merely a memory-saving or efficiency strategy, but part of a decoupled MaxEnt RL optimization scheme that restricts RL gradients to the stochastic tail.

Although the optimization is truncated at step $k_c$, the resulting training objective remains well defined. Specifically, the intermediate state $\mathbf{u}_{k_c}$ is treated as a fixed anchor when optimizing the stochastic tail, and the SAC objective therefore backpropagates only through the final $L$ denoising steps. Equivalently, the tail parameters are optimized with respect to a truncated surrogate objective of the form
\begin{equation}
    \mathcal{L}_{\text{trunc}}(\theta)
    =
    \mathbb{E}_{\mathbf{s}, \mathbf{u}_0, \boldsymbol{\varepsilon}}
    \left[
    \alpha \log \tilde{\pi}(\mathbf{u}_{k_c:K}\mid \mathbf{s}, \mathbf{u}_{k_c})
    -
    Q(\mathbf{s}, \mathbf{u}_K)
    \right],
    \label{eq:trunc_obj}
\end{equation}
where gradients are taken only through the tail dynamics from step $k_c$ to step $K$.

This formulation implies that the RL update concentrates on optimizing the stochastic tail refinement conditioned on the current transport endpoint. By confining the backward pass to the short stochastic tail ($L \ll K$), this mechanism substantially alleviates the vanishing and exploding gradient problems associated with long-horizon BPTT, thereby promoting more robust and stable training dynamics.

However, gradient truncation introduces a challenge: the prefix parameters receive no update signal from the RL objective. To address this, our framework adopts a decoupled learning strategy. While $\mathcal{L}_{\text{trunc}}$ shapes the tail for value maximization, the prefix is trained separately via flow straightening regularization (detailed in Sec. \ref{sec:fm_regularizer}). This auxiliary task not only provides the necessary supervision for the prefix but also enforces linear transport characteristics, enabling efficient few-step or even one-step sampling. This synergy ensures global policy improvement while circumventing the drawbacks of deep BPTT.

\subsection{Flow Straightening Regularization}
\label{sec:fm_regularizer}

The truncated gradient estimator proposed in Sec. \ref{sec:truncated_grad}, while stabilizing optimization, severs the RL update signal to the prefix ODE parameters. Left unconstrained, the prefix vector field may degenerate or yield highly curved trajectories, which not only violates the ``linear flow assumption'' in Sec. \ref{sec:approx_entropy} but also exacerbates numerical integration errors, rendering few-step inference infeasible.

To bridge this gap, we introduce Flow Straightening Regularization, which provides explicit supervision for the prefix while encouraging near-linear transport trajectories.

To implement this regularization, we first employ a self-distillation strategy to construct a regression target. For a given state $\mathbf{s}$ and base noise $\mathbf{u}_0$, we perform a full rollout using the current policy. To obtain a deterministic guidance signal, we disable the noise injection in the tail (i.e., set $\boldsymbol{\varepsilon}_k=\mathbf{0}$) during this rollout. This yields a trajectory endpoint $\mathbf{u}_{\text{tg}}$ that serves as a deterministic realization of the current policy. We subsequently utilize this endpoint as the fixed reference target to guide the flow straightening process.

With the source $\mathbf{u}_0$ and target $\mathbf{u}_{\text{tg}}$ fixed, the theory of Rectified Flow \cite{liu2023rectifiedflow} establishes that the optimal transport path connecting them is the straight line $\mathbf{x}_t = t \mathbf{u}_{\text{tg}} + (1-t) \mathbf{u}_0$, corresponding to a constant ideal velocity field $\mathbf{v}^* = \mathbf{u}_{\text{tg}} - \mathbf{u}_0$. Accordingly, we minimize the mean squared error between the current velocity field and this ideal direction over the prefix interval $t \in [0, \tau_{\text{cut}}]$:
\begin{equation}
    \mathcal{L}_{\text{fm}}(\theta) = \mathbb{E}_{\mathbf{s}, \mathbf{u}_0, t \sim \mathcal{U}[0, \tau_{\text{cut}}]} \left[ \left\| \mathbf{v}_\theta^{\text{ode}}(\mathbf{s}, \mathbf{x}_t, t) - (\mathbf{u}_{\text{tg}} - \mathbf{u}_0) \right\|^2 \right].
    \label{eq:lfm}
\end{equation}
This objective effectively performs online Rectified Flow training, continuously adapting the flow field to approximate the straightened paths induced by the evolving RL policy.

Mathematically, minimizing the mean squared error in Eq. \eqref{eq:lfm} drives the learned velocity field $\mathbf{v}_\theta^{\text{ode}}$ towards the conditional expectation $\mathbb{E}[\mathbf{u}_{\text{tg}} - \mathbf{u}_0 | \mathbf{x}_t]$. Since the target trajectory is constructed via linear interpolation, the optimal velocity field tends to be a constant vector independent of time. 
This geometric property yields two critical advantages. First, as discussed in Sec. \ref{sec:approx_entropy}, a constant field implies vanishing divergence, thereby minimizing the approximation error of our surrogate log-likelihood. Second, it significantly enhances sampling efficiency. Since linear trajectories facilitate accurate numerical integration with minimal discretization error, this property allows TRFP to maintain high performance when reducing the sampling horizon, extending the model's capability to few-step and even one-step sampling.

Finally, the total optimization objective of TRFP combines the truncated RL loss $\mathcal{L}_{\text{trunc}}$ and the straightening regularization $\mathcal{L}_{\text{fm}}$:
\begin{equation}
    \mathcal{L}_{\text{total}}(\theta) = \mathcal{L}_{\text{trunc}}(\theta) + \lambda_{\text{fm}} \mathcal{L}_{\text{fm}}(\theta).
    \label{eq:actor_total}
\end{equation}
This combination realizes a decoupled optimization paradigm: $\mathcal{L}_{\text{trunc}}$ drives exploration and value maximization, while $\mathcal{L}_{\text{fm}}$ regularizes the prefix transport to improve trajectory straightness and sampling efficiency.

\begin{algorithm}[t]
\caption{Truncated Rectified Flow Policy (TRFP)}
\label{alg:trfp}
\begin{algorithmic}[1]
\STATE \textbf{Initialize:} Policy $\theta$, Critics $\psi_{1,2}$, Target Critics $\bar{\psi}_{1,2}$, Temperature $\alpha$, and Replay Buffer $\mathcal{D}$.
\FOR{each environment step}
    \STATE Sample action $\mathbf{a}_t \sim \pi_{\theta}(\cdot|\mathbf{s}_t)$ via the hybrid sampler (Sec. \ref{sec:hybrid_policy}).
    \STATE Execute $\mathbf{a}_t$, store transition $(\mathbf{s}_t, \mathbf{a}_t, r_t, \mathbf{s}_{t+1})$ in $\mathcal{D}$.
    
    \FOR{each update step}
        \STATE Sample batch $\mathcal{B} = \{(\mathbf{s}, \mathbf{a}, r, \mathbf{s}')\} \sim \mathcal{D}$.
        
        \STATE Update critics $\psi_{1,2}$ by minimizing $\mathcal{L}_Q(\psi)$ \eqref{eq:critic_loss}, using surrogate $\log \tilde{\pi}$ \eqref{eq:surrogate_logp} for targets.
        \STATE Soft-update target critics $\bar{\psi}_{1,2} \leftarrow (1-\tau)\bar{\psi}_{1,2} + \tau \psi_{1,2}$.
        
        \STATE Compute truncated RL loss $\mathcal{L}_{\text{trunc}}(\theta)$ via \eqref{eq:trunc_obj}.
        \STATE Compute flow straightening loss $\mathcal{L}_{\text{fm}}(\theta)$ via \eqref{eq:lfm}.
        \STATE Update policy $\theta$ by minimizing $\mathcal{L}_{\text{total}}$ \eqref{eq:actor_total}.
        
        \STATE Update temperature $\alpha$ by minimizing $\mathcal{L}_{\alpha}$ \eqref{eq:alpha_loss}, using surrogate $\log \tilde{\pi}$ \eqref{eq:surrogate_logp}.
    \ENDFOR
\ENDFOR
\end{algorithmic}
\end{algorithm}

\subsection{Implementation Details}
\label{sec:implementation_details}

While the training paradigm of TRFP is summarized in Algorithm \ref{alg:trfp}, we adopt specific strategies during evaluation to balance computational efficiency and decision quality.

\textbf{Accelerated Few-Step Sampling.}
While the hybrid stochastic architecture is essential for exploration during training, we omit the stochastic tail during evaluation and generate actions using only deterministic transport. 
Benefiting from the Flow Straightening Regularization introduced in Sec. \ref{sec:fm_regularizer}, the learned velocity field exhibits highly linear characteristics. This geometric property allows us to substantially reduce the number of numerical integration steps during inference without significantly compromising generation fidelity. 
Empirically, we observe that a simple one-step integration is often sufficient to yield high-precision action approximations. This strategy minimizes inference complexity, enabling TRFP to meet the requirements of real-time control.

\textbf{Q-Guided Action Selection.} To further exploit the generative capabilities of TRFP, we adopt the widely used value-based Monte Carlo selection strategy \cite{lu2025what}. Given that TRFP models a complex multimodal action distribution, a single sample may not always capture the global optimum. Consequently, we utilize the trained critic networks to evaluate $N$ action candidates $\{\mathbf{a}_i\}_{i=1}^N$ in parallel and execute the one with the highest predicted Q-value:
\begin{equation}
    \mathbf{a}^* = \arg\max_{\mathbf{a}_i} Q_{\psi}(\mathbf{s}, \mathbf{a}_i).
\end{equation}
This mechanism acts as an implicit rejection sampling process, effectively filtering out low-value outliers and significantly enhancing policy robustness and cumulative returns under computational constraints.

\section{Experiments}
\label{experiments}
In this section, we present the experimental results of TRFP. We first analyze its multimodal behavior in a toy multigoal environment, and then evaluate it on 10 MuJoCo continuous-control benchmarks against SAC, TD3, SDAC, and MaxEntDP, followed by ablation studies on the contribution of each component.
\subsection{Multimodality Analysis in a Toy Multigoal Environment}
\label{sec:multimodality_analysis_in_a_toy_multigoal_environment}
To verify that TRFP provides a stronger capacity for multimodal policy representation, we visualize the behavior trajectories of different methods in a toy multigoal environment.
The environment contains four goals with identical reward values, symmetrically located in the four directions of the state space. For each method, we collect 20 evaluation episodes after training and overlay all trajectories.

\begin{figure}[!tbp]
    \centering
    \subfloat[TRFP\label{fig:multigoal_trfp}]{%
        \includegraphics[width=0.32\linewidth]{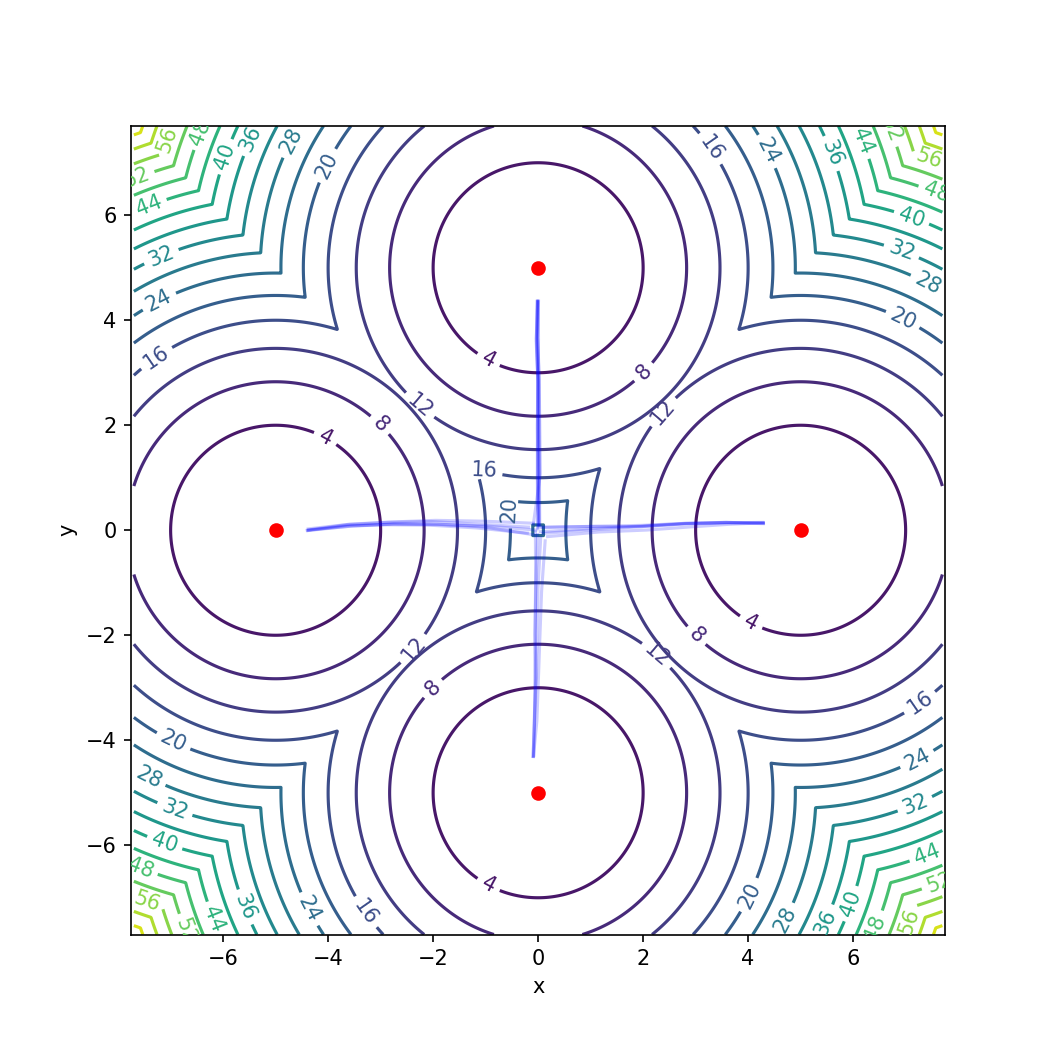}%
    }%
    \hfill
    \subfloat[SAC\label{fig:multigoal_sac}]{%
        \includegraphics[width=0.32\linewidth]{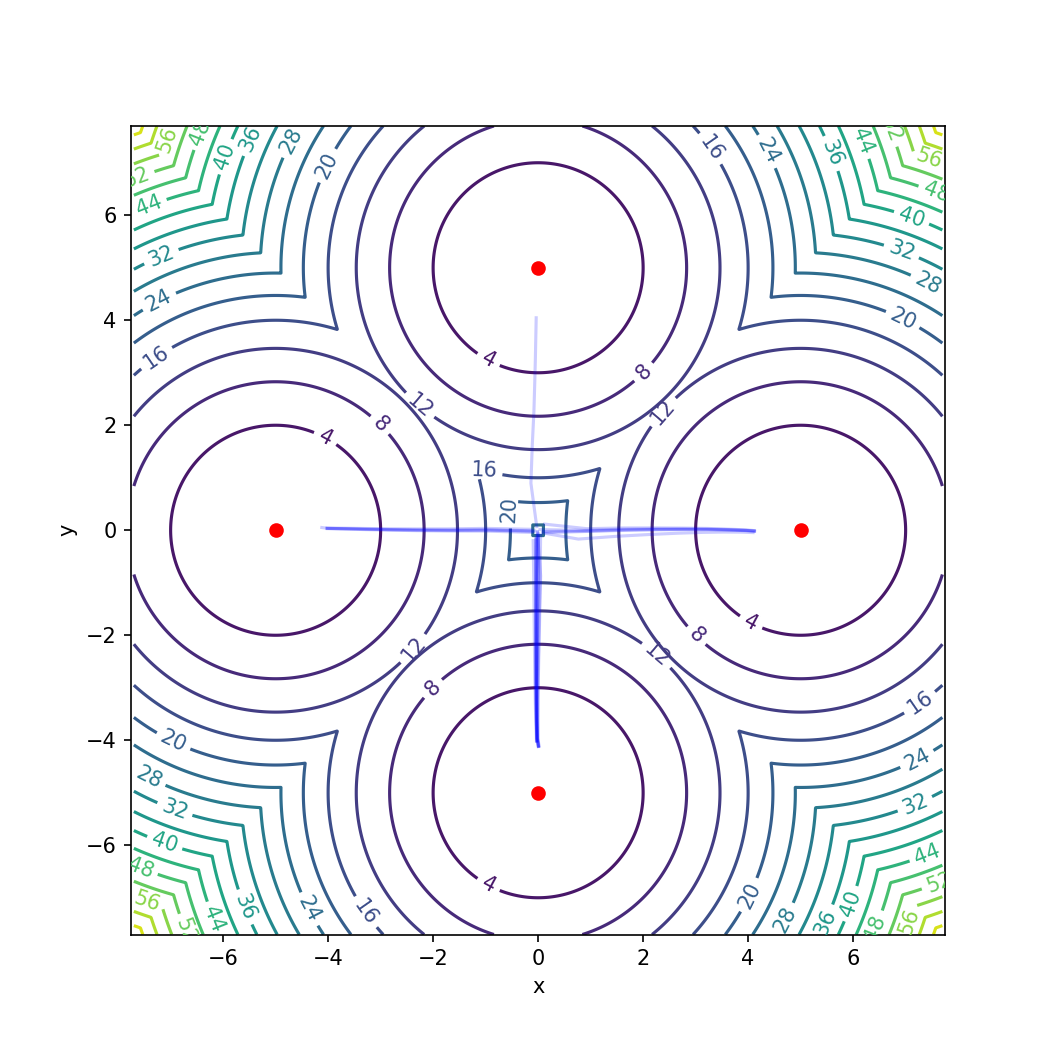}%
    }%
    \hfill
    \subfloat[TD3\label{fig:multigoal_td3}]{%
        \includegraphics[width=0.32\linewidth]{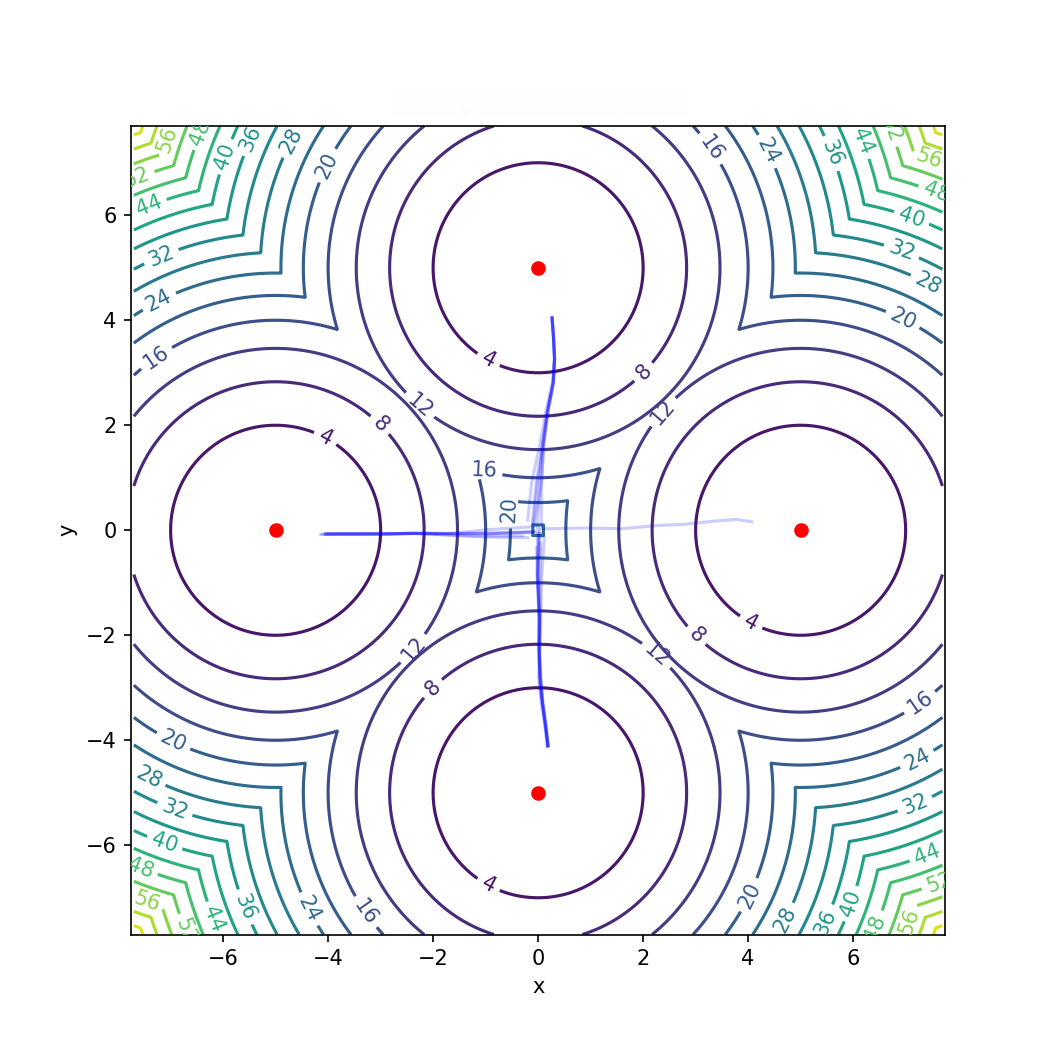}%
    }
    
    \caption{Trajectory visualization in the toy multigoal environment. The agent starts near the center and moves toward four equally rewarded goals. Each panel overlays 20 evaluation episodes collected after training.}
    \label{fig:multigoal_traj}
\end{figure}

As shown in Fig.~\ref{fig:multigoal_traj}, TRFP covers all four goal directions with more balanced occupancy, suggesting that it better preserves multimodal behaviors under equal-reward multi-goal settings. In contrast, SAC and TD3 are biased toward only a subset of the goals, covering only part of the available modes. This observation highlights the advantage of TRFP in representing complex policy distributions.

\subsection{Comparative Evaluation}
\label{sec:comparative_evaluation}
In this subsection, we evaluate TRFP on 10 MuJoCo benchmark tasks and compare it with SDAC, MaxEntDP, SAC, and TD3. Among these baselines, SAC and TD3 represent classical off-policy actor-critic methods, while SDAC and MaxEntDP represent recent diffusion-based  MaxEnt RL approaches.

By default, TRFP is trained under the standard sampling configuration with $K=4$ and $L=1$. 
In addition, we report a one-step evaluation protocol, described in Sec.~\ref{sec:implementation_details}, to assess its applicability in low-latency deployment. 
For evaluation, we omit the stochastic tail and use deterministic transport only: the standard evaluation protocol uses 4 deterministic steps, whereas the one-step protocol uses a single deterministic step.
Each method is trained independently with five random seeds. The complete hyperparameter settings are provided in Appendix Table~\ref{tab:hyperparams}.

All methods are implemented based on their official codebases. For SAC, TD3, and TRFP, we use three-layer MLPs with Mish activations for both the actor and critic, while the remaining baselines follow their original implementations. We keep the shared training hyperparameters aligned whenever applicable. 
All experiments are conducted on a GPU of single NVIDIA RTX GeForce 4090 and a CPU of Intel Core i9-14900KF.

\begin{figure}[!tbp]
    \centering
    \subfloat[Humanoid-v5\label{fig:humanoid_v5}]{%
        \includegraphics[width=0.32\linewidth]{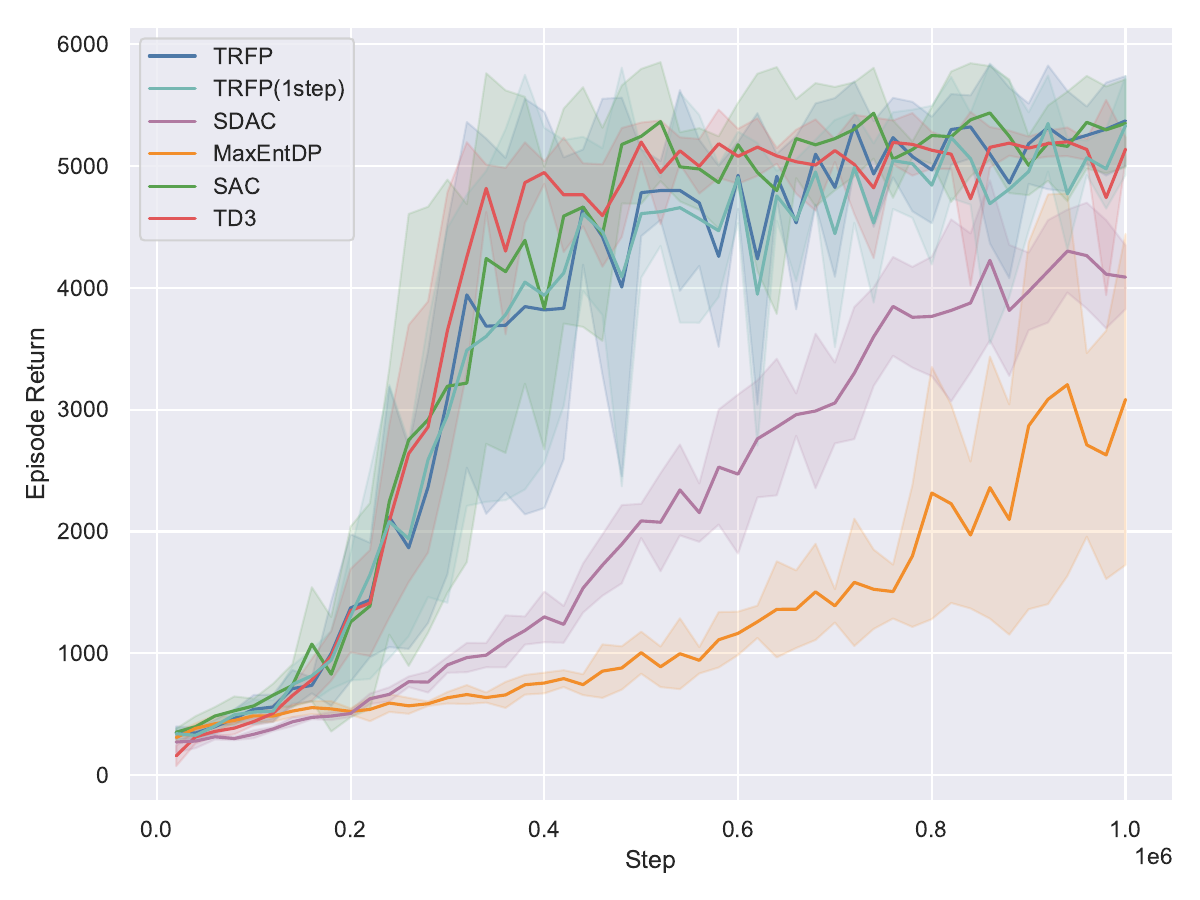}%
    }\hfill
    \subfloat[HalfCheetah-v5\label{fig:halfcheetah_v5}]{%
        \includegraphics[width=0.32\linewidth]{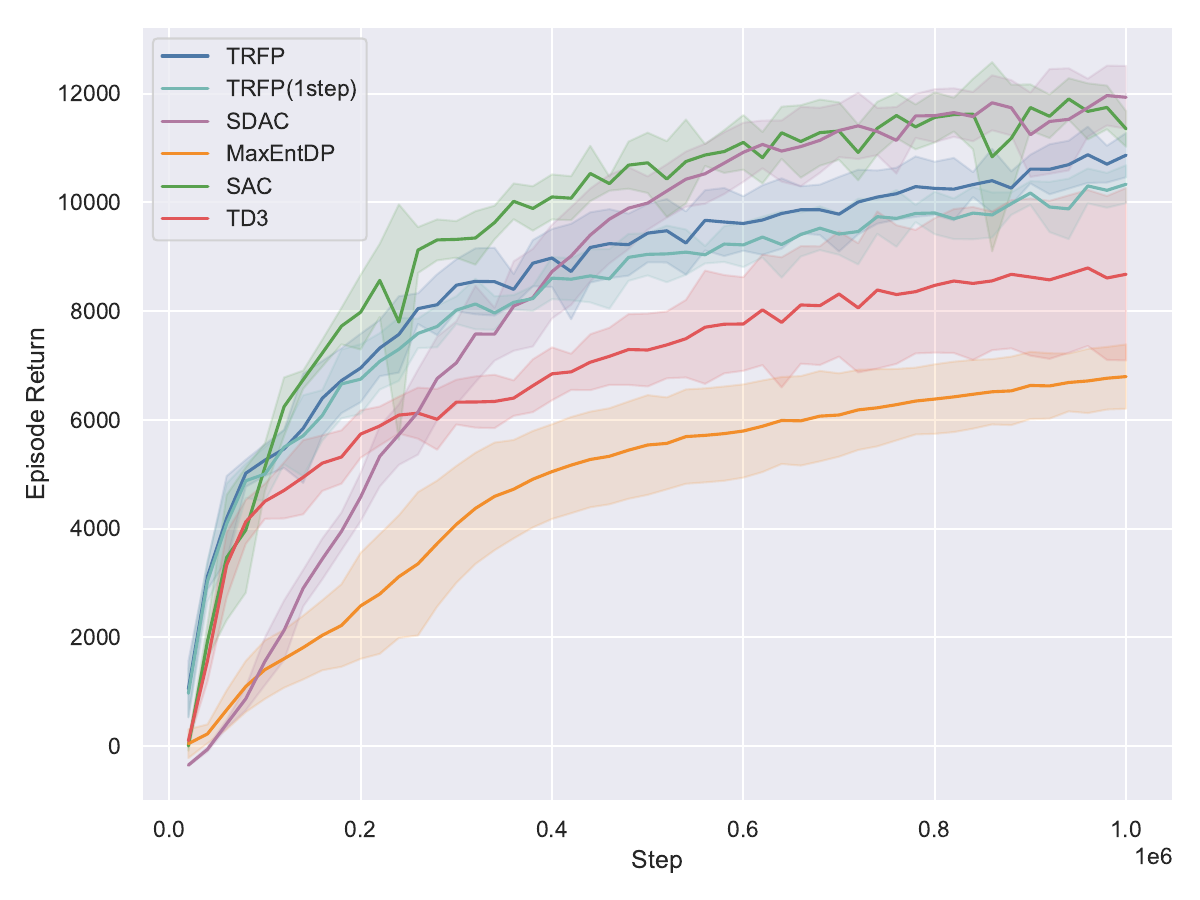}%
    }\hfill
    \subfloat[Hopper-v5\label{fig:hopper_v5}]{%
        \includegraphics[width=0.32\linewidth]{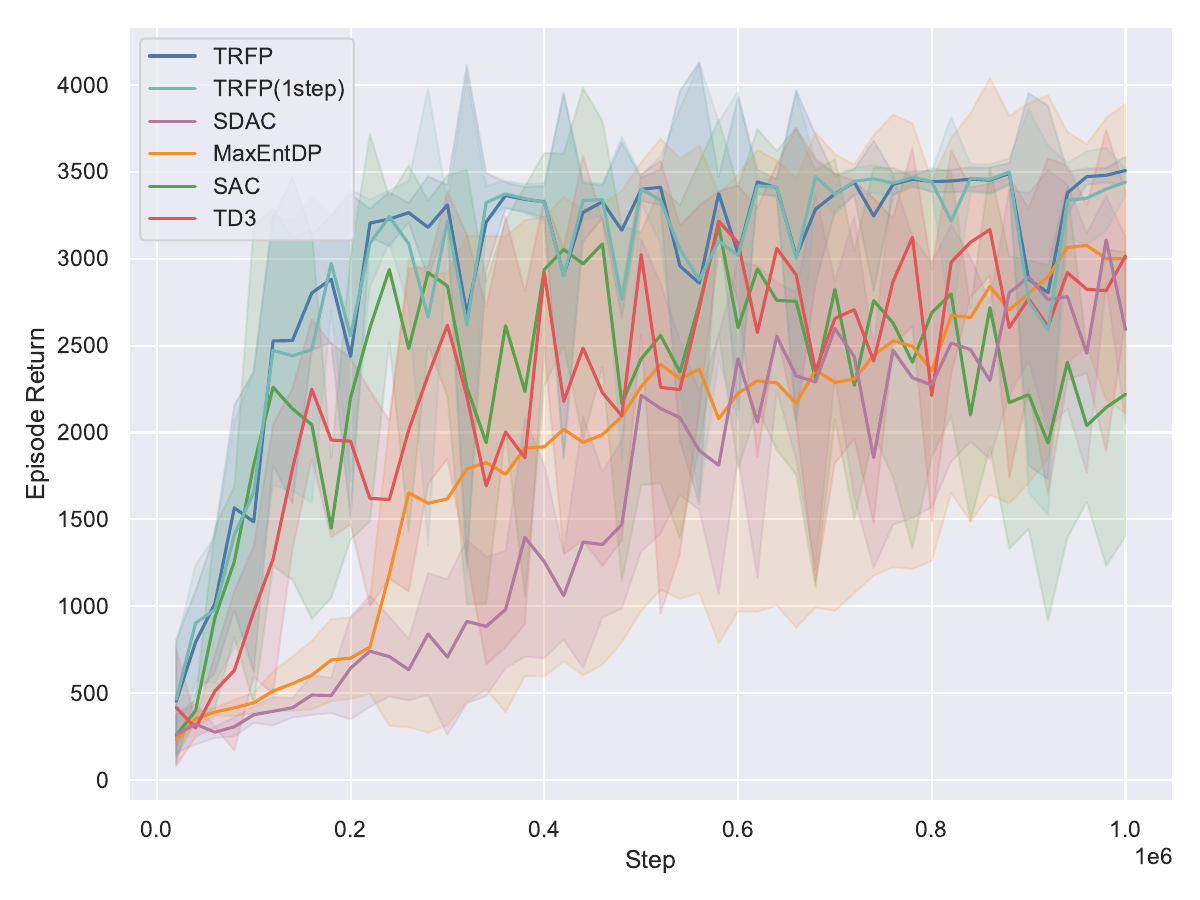}%
    }
    
    \vspace{0.6em}
    
    \subfloat[Walker2d-v5\label{fig:walker2d_v5}]{%
        \includegraphics[width=0.32\linewidth]{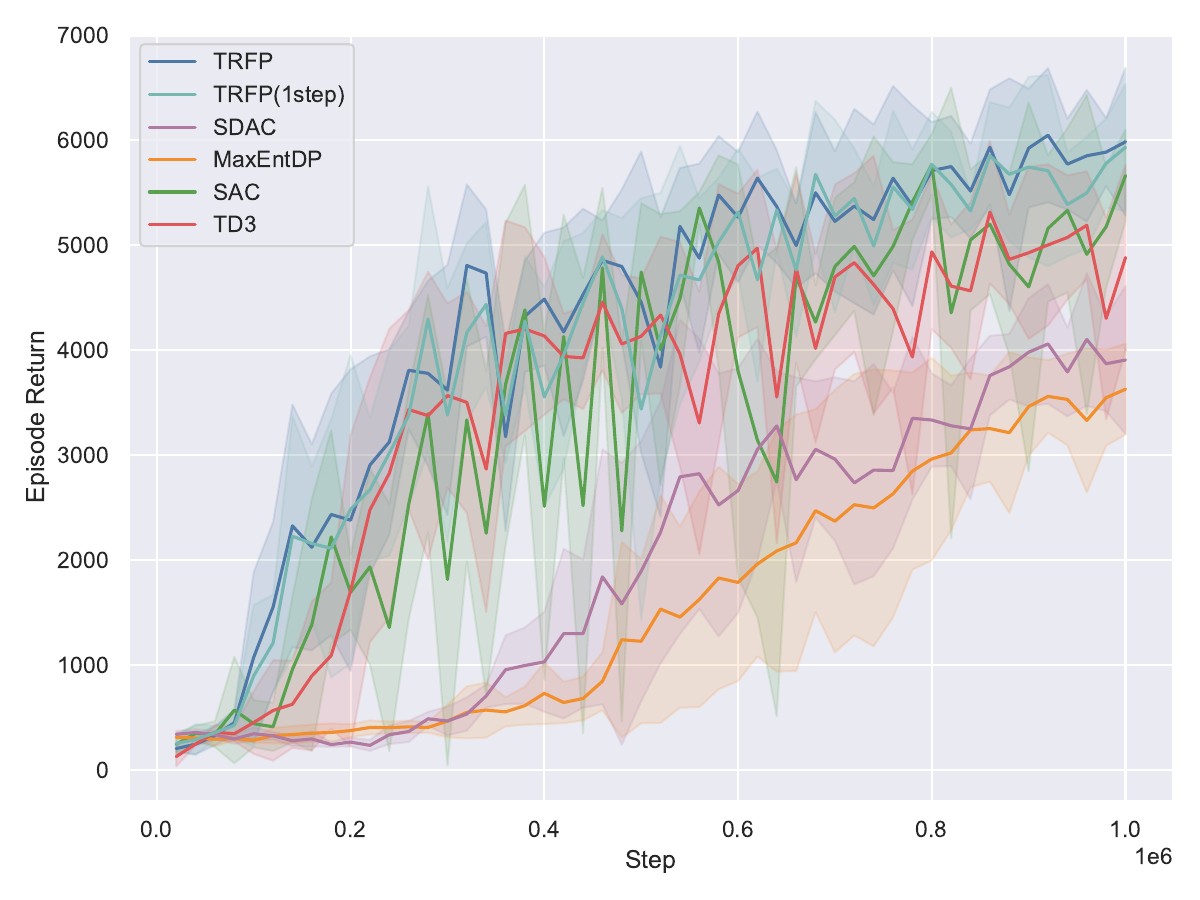}%
    }\hfill
    \subfloat[Ant-v5\label{fig:ant_v5}]{%
        \includegraphics[width=0.32\linewidth]{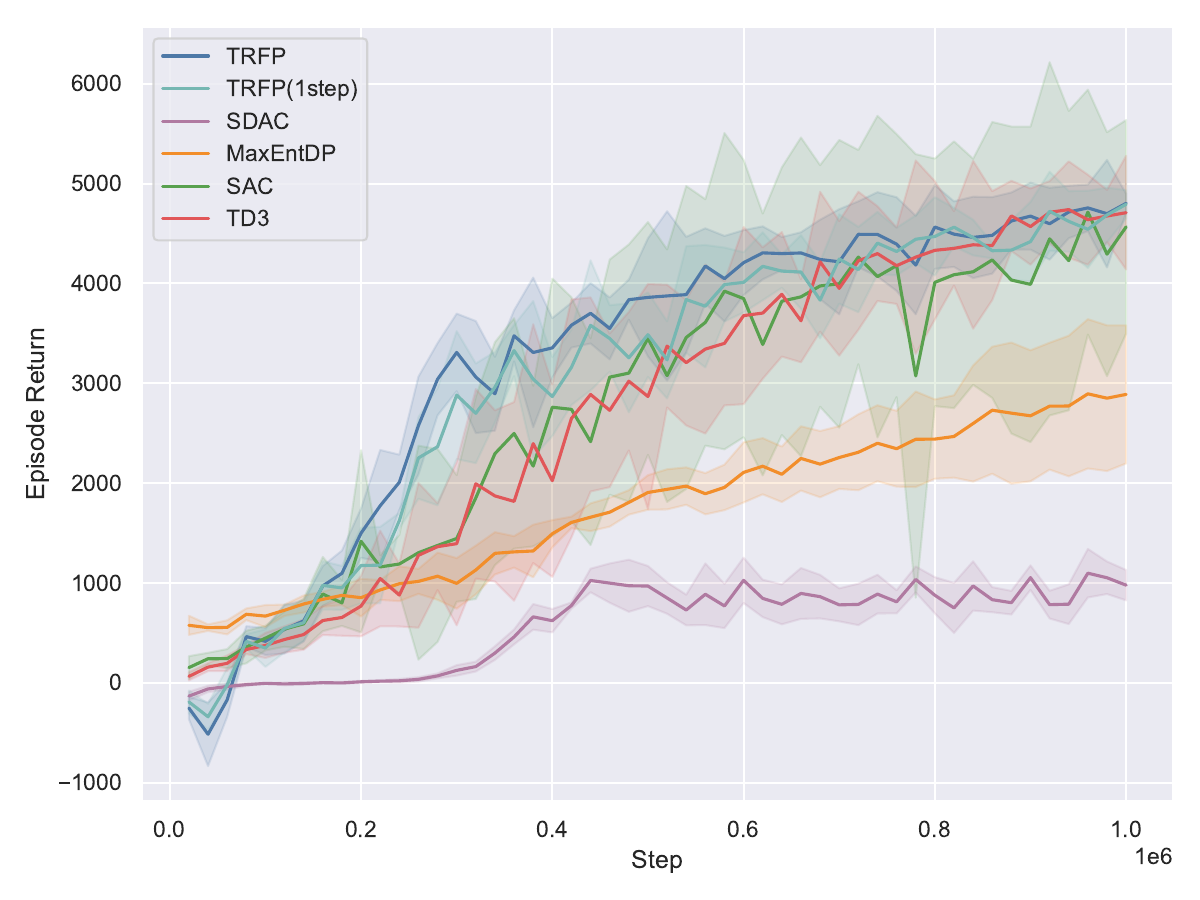}%
    }\hfill
    \subfloat[Swimmer-v5\label{fig:swimmer_v5}]{%
        \includegraphics[width=0.32\linewidth]{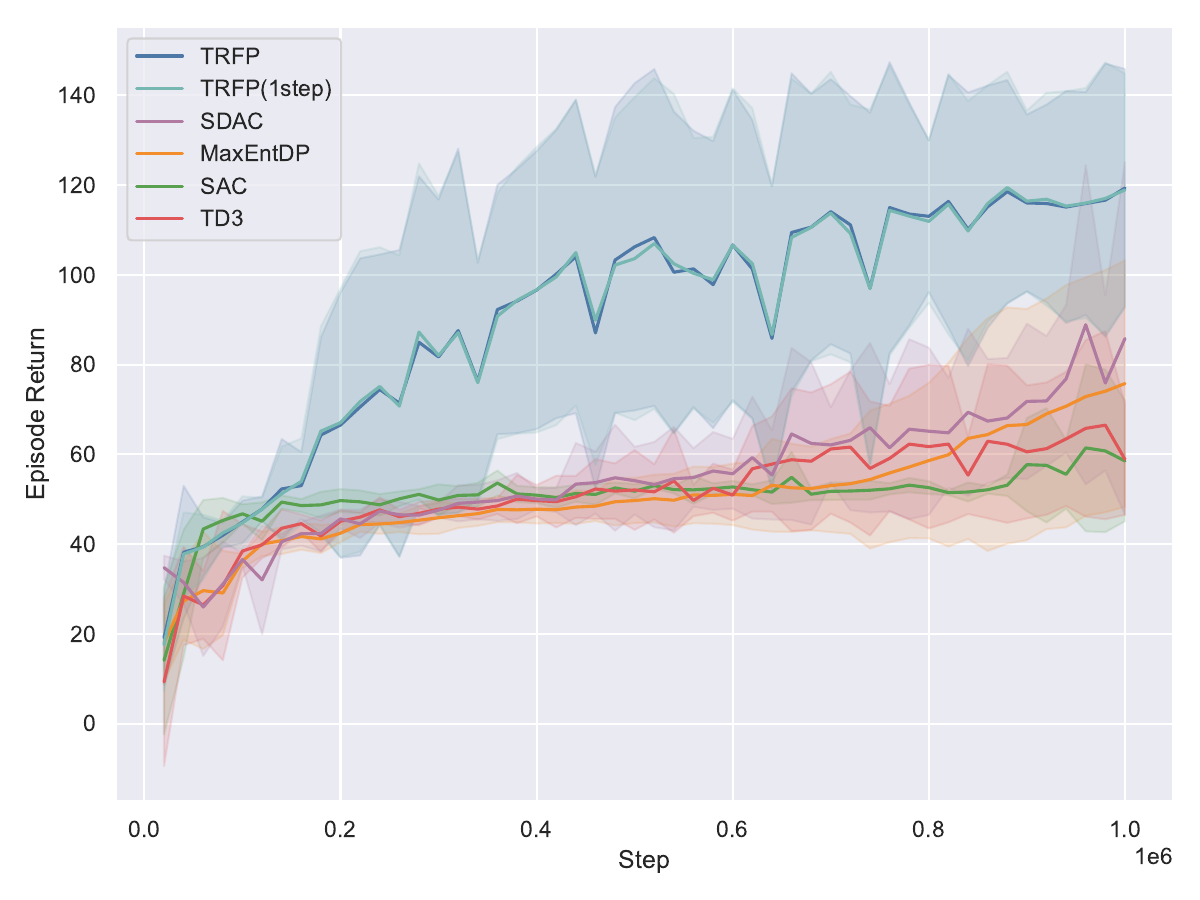}%
    }
    
    \vspace{0.6em}
    
    \subfloat[Reacher-v5\label{fig:reacher_v5}]{%
        \includegraphics[width=0.32\linewidth]{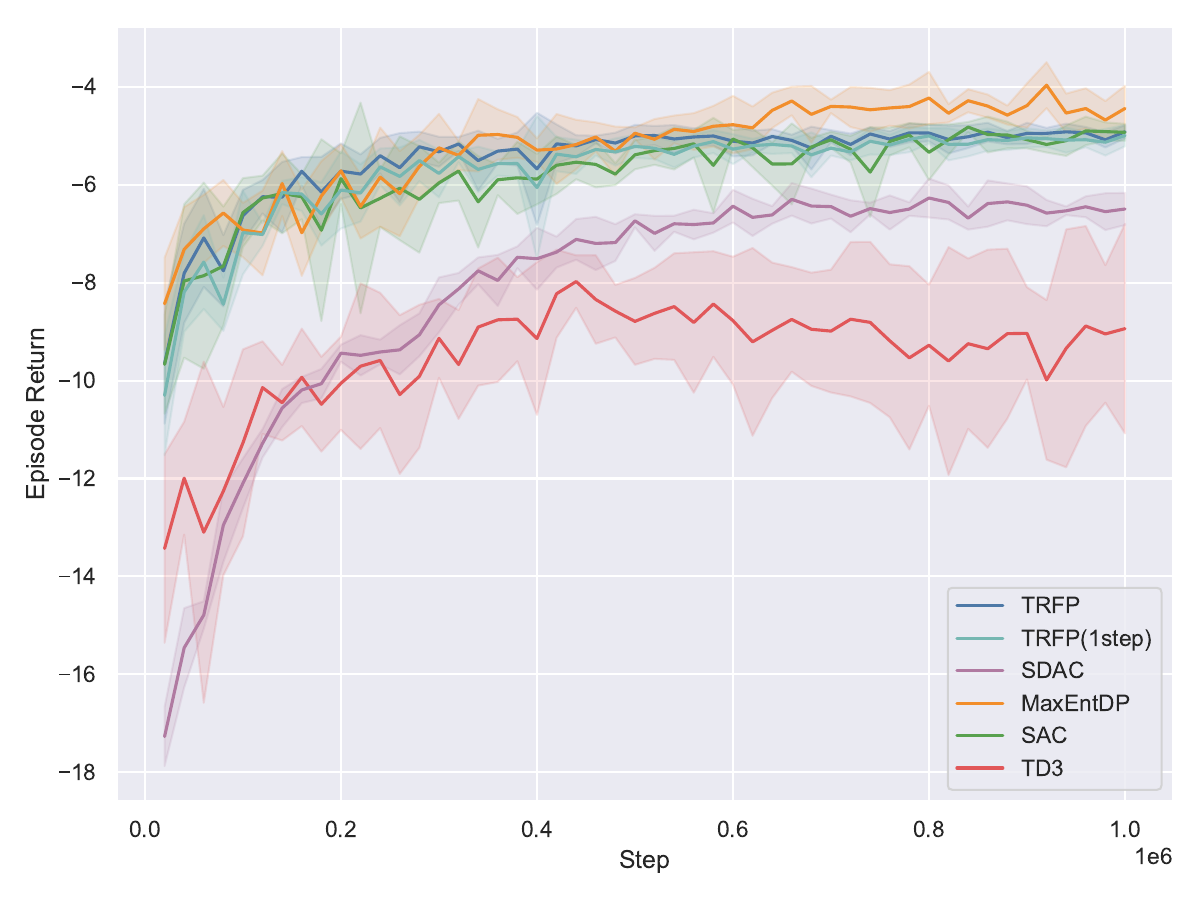}%
    }\hfill
    \subfloat[InvertedPendulum-v5\label{fig:invertedpendulum_v5}]{%
        \includegraphics[width=0.32\linewidth]{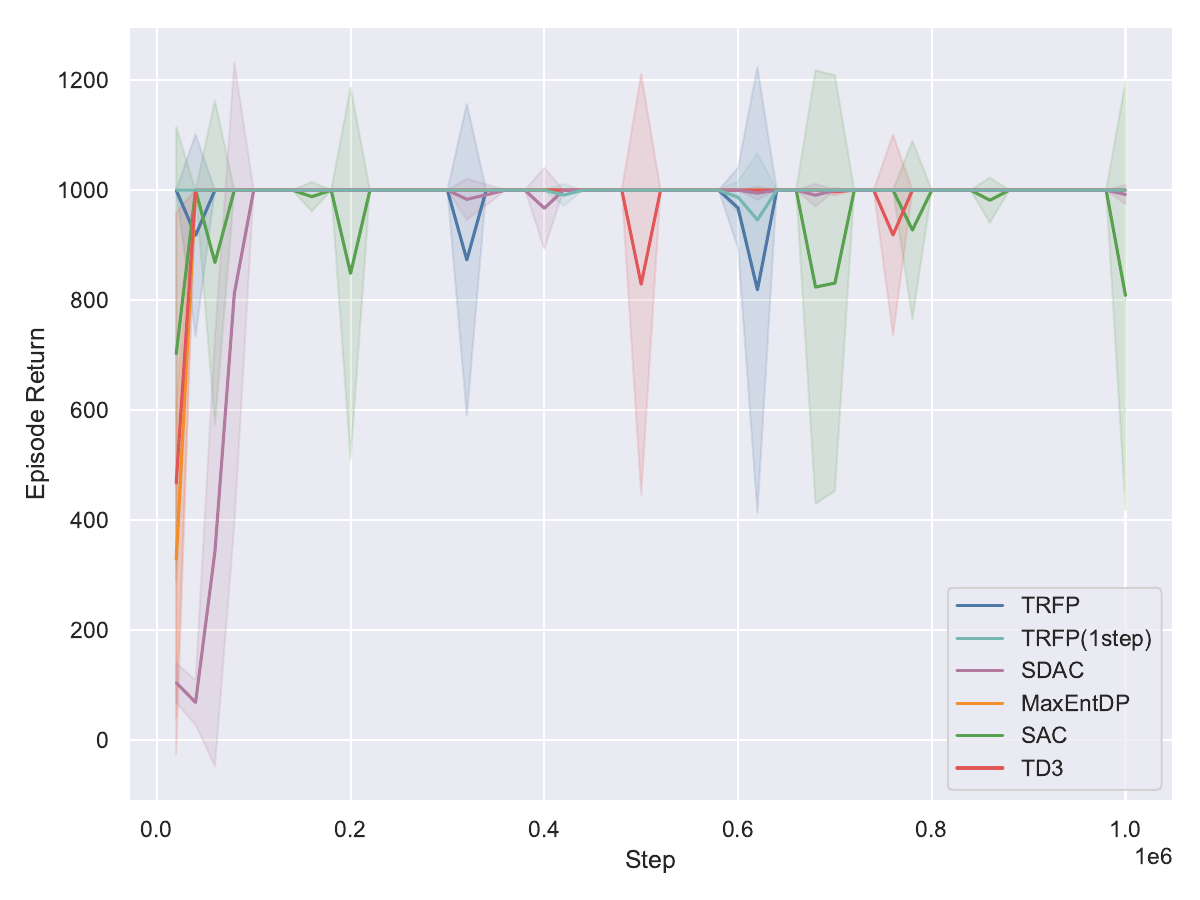}%
    }\hfill
    \subfloat[InvertedDoublePendulum-v5\label{fig:inverteddoublependulum_v5}]{%
        \includegraphics[width=0.32\linewidth]{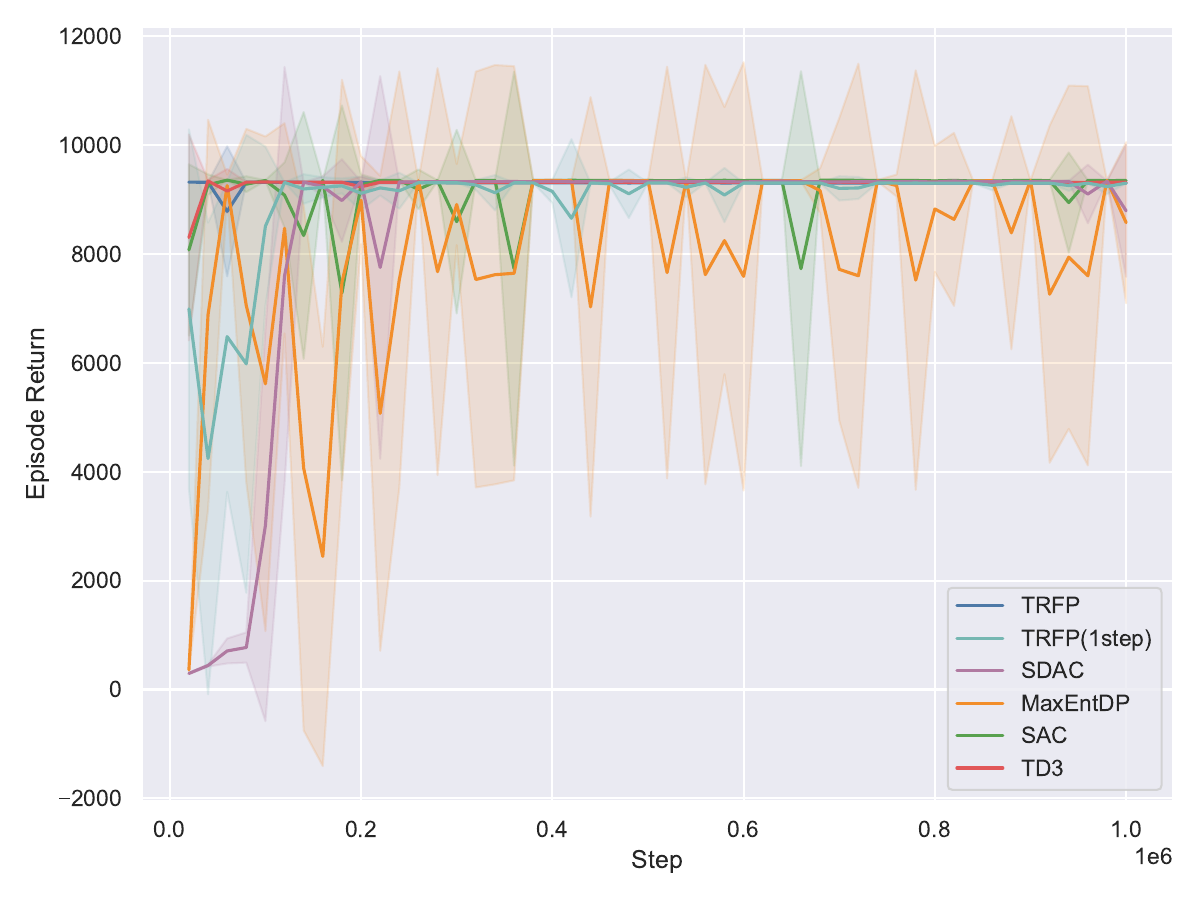}%
    }
    
    \vspace{0.6em}
    
    \subfloat[Pusher-v5\label{fig:pusher_v5}]{%
        \includegraphics[width=0.32\linewidth]{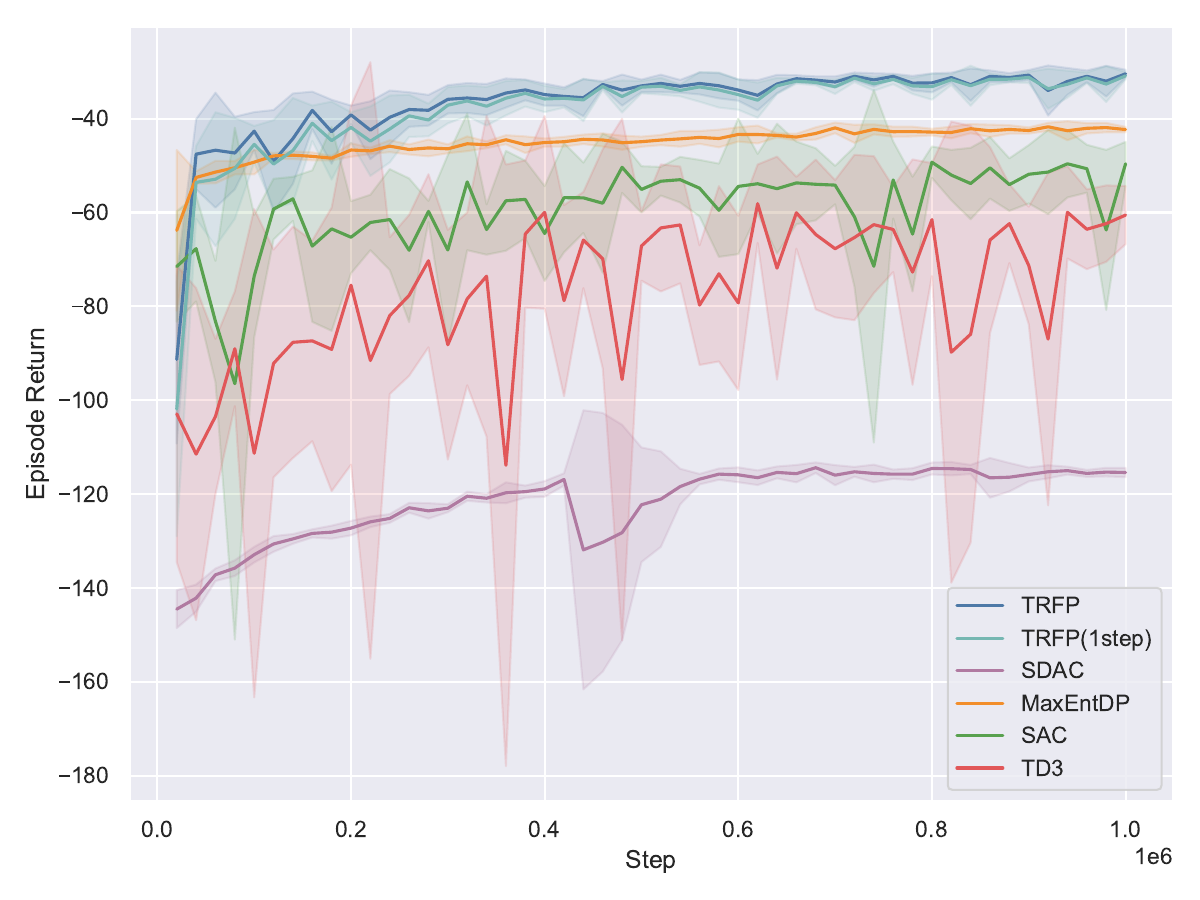}%
    }

    \caption{Learning curves on 10 MuJoCo benchmarks. Each subplot compares TRFP under standard sampling ($K=4$) and one-step sampling ($K=1$) with four baselines. The solid lines denote the mean episode return over five random seeds, and the shaded regions represent one standard deviation.}
    \label{fig:mujoco_curves}
\end{figure}

Fig.~\ref{fig:mujoco_curves} presents the learning curves on 10 MuJoCo-v5 benchmark tasks for TRFP, TRFP(one-step), SDAC, MaxEntDP, SAC, and TD3. Under standard sampling ($K=4$), TRFP achieves strong and consistent performance across the benchmark suite, outperforming the baselines on several environments while remaining competitive on the others. This indicates that the proposed framework combines expressive policy modeling with strong performance on standard continuous-control tasks.

More importantly, when inference is further compressed to one-step sampling ($K=1$), TRFP(one-step) remains close to standard TRFP on most tasks and still compares favorably with all baselines across the benchmark suite. 
This suggests that the proposed method combines expressive policy representation with strong performance on standard continuous-control tasks. 
For a compact comparison, Table~\ref{tab:final_return} summarizes the final returns on all MuJoCo-v5 benchmarks together with the per-action inference cost measured by the number of function evaluations (NFE).

\begin{table*}[t]
\centering
\caption{Final return comparison on 10 MuJoCo-v5 benchmarks. Results are reported as mean $\pm$ standard deviation. Best results are shown in \textbf{bold}, and second-best distinct results are \underline{underlined}. NFE denotes the number of function evaluations per action. For iterative generative policies, expressions of the form $a \times b$ indicate $a$ rollout steps with $b$ sampled action candidates.}

\label{tab:final_return}
\setlength{\tabcolsep}{4pt}
\scriptsize

\begin{tabular}{lcccccc}
\toprule
Method & NFE & Ant-v5 & HalfCheetah-v5 & Hopper-v5 & Humanoid-v5 & Inverted2Pendulum-v5 \\
\midrule
TD3 & 1 & 4708.5 $\pm$ 569.1 & 8677.1 $\pm$ 1585.6 & 3015.6 $\pm$ 370.4 & 5138.6 $\pm$ 96.7 & \underline{9319.4 $\pm$ 0.4} \\
SAC & 1 & 4566.4 $\pm$ 1069.0 & \underline{11349.7 $\pm$ 326.5} & 2220.6 $\pm$ 815.9 & \underline{5350.5 $\pm$ 361.5} & \textbf{9352.0 $\pm$ 6.4} \\
MaxEntDP & $20 \times 10$ & 2888.6 $\pm$ 691.9 & 6794.8 $\pm$ 591.4 & 3000.6 $\pm$ 888.7 & 3084.2 $\pm$ 1358.8 & 8578.2 $\pm$ 1478.4 \\
SDAC & $20 \times 32$ & 977.5 $\pm$ 152.2 & \textbf{11931.2 $\pm$ 578.4} & 2591.3 $\pm$ 539.3 & 4087.9 $\pm$ 262.0 & 8794.3 $\pm$ 1210.9 \\
TRFP(ours) & $4 \times 4$ & \textbf{4802.1 $\pm$ 95.6} & 10867.5 $\pm$ 406.4 & \textbf{3507.5 $\pm$ 79.6} & \textbf{5370.8 $\pm$ 367.6} & 9304.1 $\pm$ 2.6 \\
TRFP(one-step) & $1 \times 4$ & \underline{4792.8 $\pm$ 142.3} & 10332.7 $\pm$ 348.1 & \underline{3440.6 $\pm$ 75.6} & 5332.5 $\pm$ 406.5 & 9304.1 $\pm$ 2.6 \\
\bottomrule

\noalign{\vspace{1em}}

\toprule
Method & NFE & InvertedPendulum-v5 & Pusher-v5 & Reacher-v5 & Swimmer-v5 & Walker2d-v5 \\
\midrule
TD3 & 1 & \textbf{1000.0 $\pm$ 0.0} & -60.5 $\pm$ 6.2 & -8.9 $\pm$ 2.1 & 58.9 $\pm$ 12.4 & 4883.4 $\pm$ 882.4 \\
SAC & 1 & 808.1 $\pm$ 390.0 & -49.6 $\pm$ 4.7 & -4.9 $\pm$ 0.2 & 58.5 $\pm$ 13.4 & 5665.4 $\pm$ 437.3 \\
MaxEntDP & $20 \times 10$ & \textbf{1000.0 $\pm$ 0.0} & -42.3 $\pm$ 0.6 & \textbf{-4.4 $\pm$ 0.5} & 75.8 $\pm$ 27.5 & 3628.5 $\pm$ 433.0 \\
SDAC & $20 \times 32$ & \underline{992.1 $\pm$ 17.7} & -115.4 $\pm$ 1.0 & -6.5 $\pm$ 0.3 & 85.8 $\pm$ 39.3 & 3906.5 $\pm$ 706.2 \\
TRFP(ours) & $4 \times 4$ & \textbf{1000.0 $\pm$ 0.0} & \textbf{-30.4 $\pm$ 0.9} & \underline{-4.9 $\pm$ 0.1} & \textbf{119.3 $\pm$ 26.5} & \textbf{5986.9 $\pm$ 706.4} \\
TRFP(one-step) & $1 \times 4$ & \textbf{1000.0 $\pm$ 0.0} & \underline{-30.9 $\pm$ 0.8} & -5.0 $\pm$ 0.2 & \underline{118.9 $\pm$ 25.9} & \underline{5933.9 $\pm$ 604.1} \\
\bottomrule
\end{tabular}
\end{table*}

\subsection{Ablation Studies}
\label{sec:ablation_studies}

\begin{figure}[!tbp]
    \centering
    \subfloat[HalfCheetah-v5\label{fig:ablation_wolfm}]{%
        \includegraphics[width=0.32\linewidth]{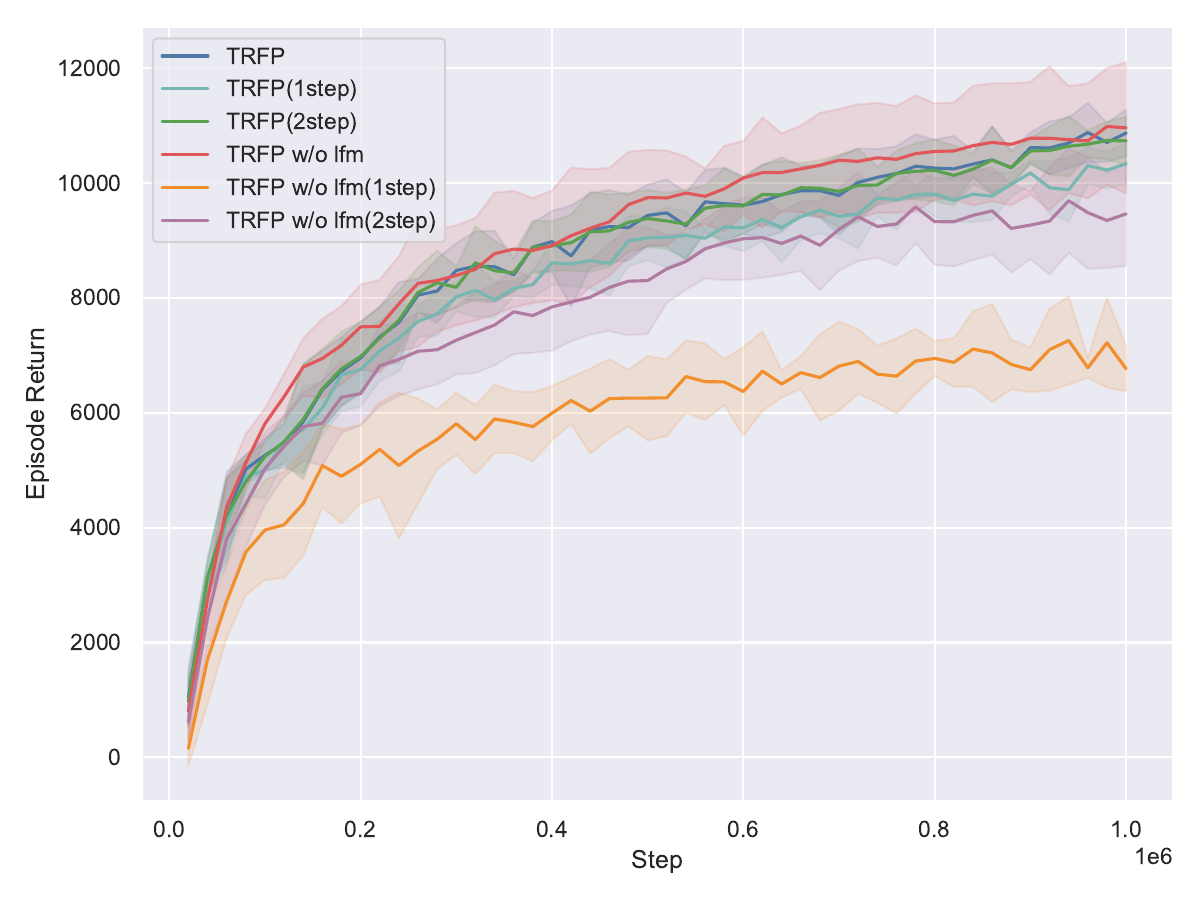}%
    }\hfill
    \subfloat[HalfCheetah-v5\label{fig:ablation_womaxq}]{%
        \includegraphics[width=0.32\linewidth]{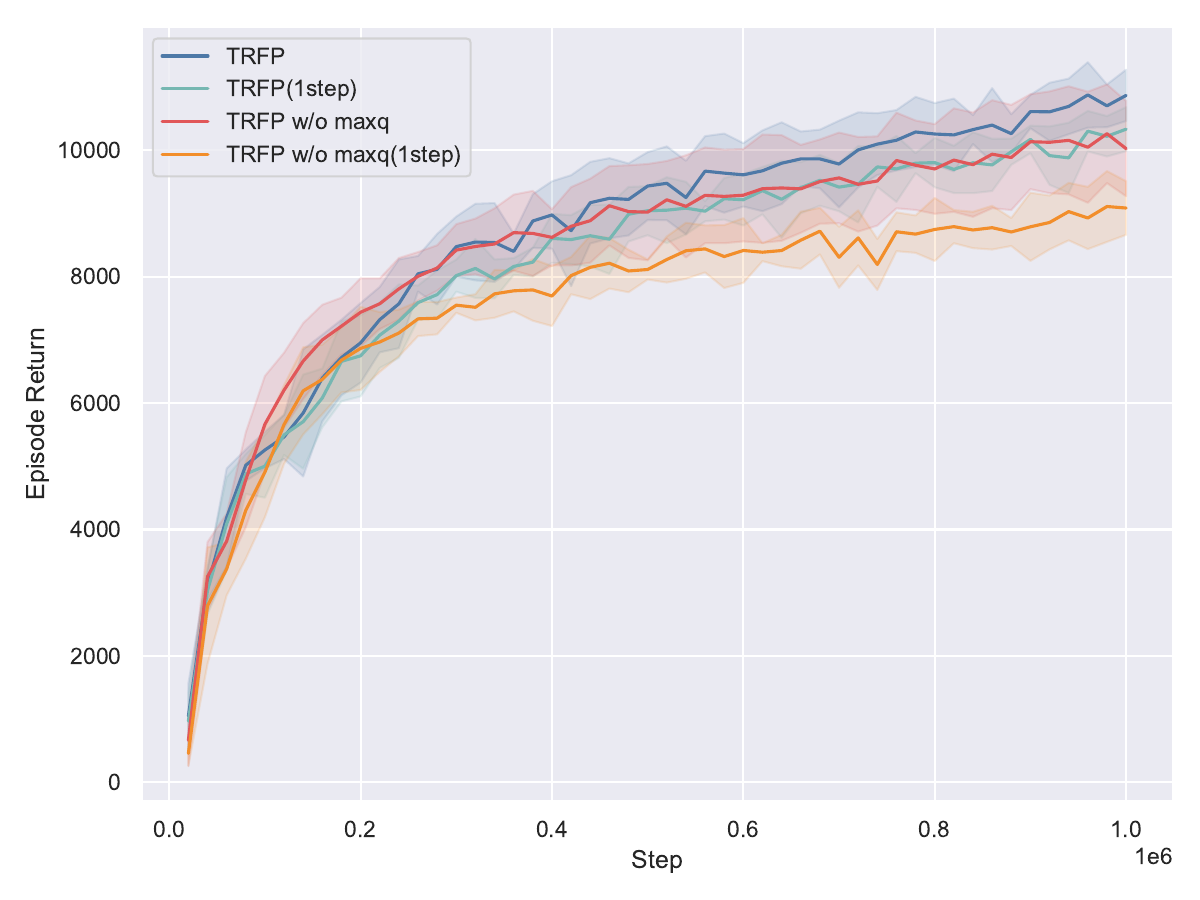}%
    }\hfill
    \subfloat[Humanoid-v5\label{fig:ablation_wotail}]{%
        \includegraphics[width=0.32\linewidth]{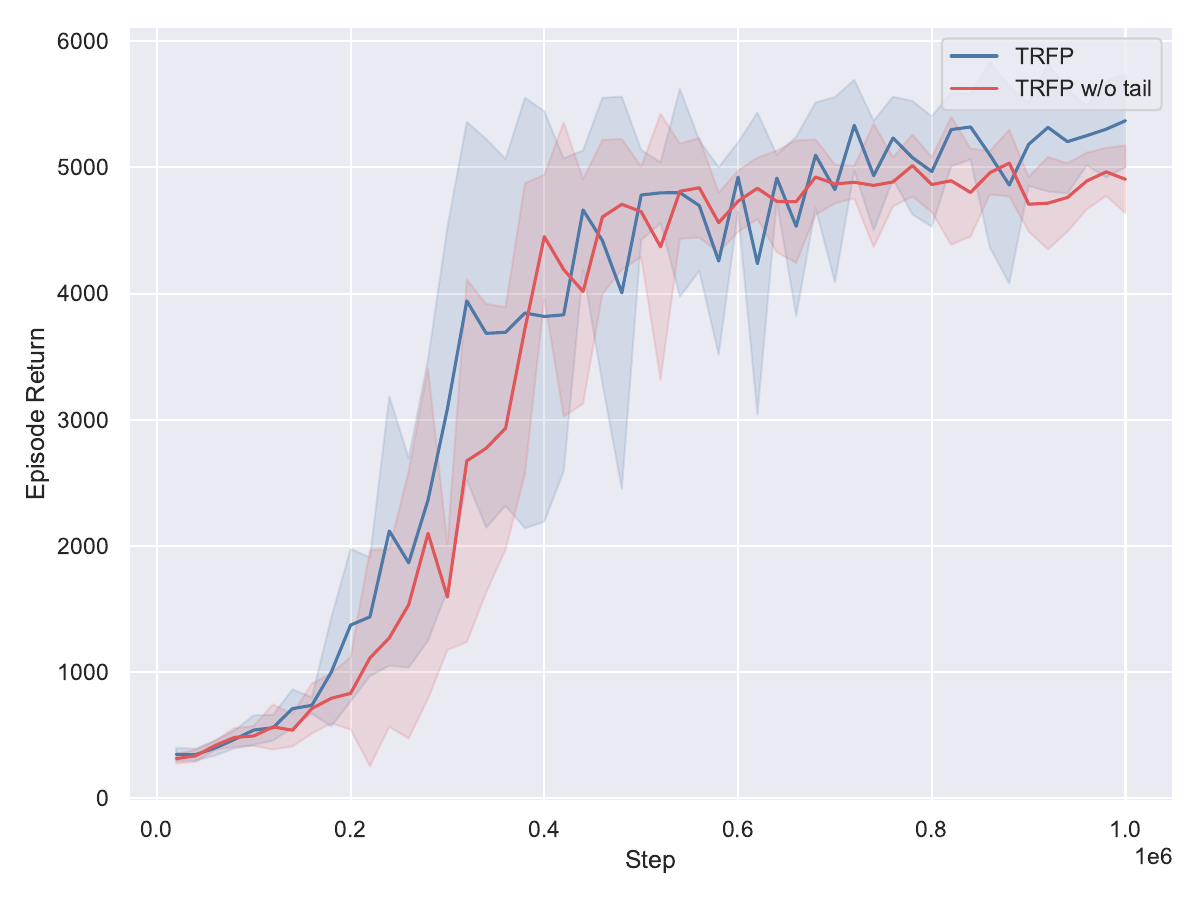}%
    }
    
    \caption{Ablation results of TRFP. From left to right: the effect of removing flow straightening regularization, the effect of removing Q-guided action selection, and the effect of removing the stochastic tail. The first two panels are evaluated on HalfCheetah-v5, and the third panel is evaluated on Humanoid-v5. Solid lines denote the mean episode return over five random seeds, and shaded regions represent one standard deviation.}
    \label{fig:ablation_main}
\end{figure}

In this subsection, we present targeted ablations on flow straightening regularization, Q-guided action selection, and the stochastic tail.

\textbf{Flow Straightening Regularization.}
Figure~\ref{fig:ablation_wolfm} presents the ablation results on HalfCheetah. Under the standard sampling setting $(K=4)$, removing $\mathcal{L}_{\text{fm}}$ yields a final return close to that of TRFP, suggesting that this regularizer is not mainly used to raise the performance ceiling under conventional multi-step sampling. 
By contrast, in the few-step setting, removing this term causes a substantial performance drop, with the return on HalfCheetah decreasing by more than 40\% under one-step sampling, indicating that its primary role is to preserve performance fidelity under fewer sampling steps.

\textbf{Q-Guided Action Selection.}
Figure~\ref{fig:ablation_womaxq} shows the corresponding ablation results on HalfCheetah. The results indicate that removing the Q-guided action selection mechanism leads to an approximately 10\% performance drop under both the standard sampling setting and the one-step setting. 

Due to the multimodal expressiveness of the policy, TRFP can generate multiple candidate actions, which enables the Q-guided action selection mechanism to utilize the critic to select higher-value actions.
The results show that this refinement mechanism mitigates suboptimal action selection induced by sampling stochasticity, thereby delivering consistent improvements across different sampling steps.

\textbf{Stochastic Tail.} 
Finally, we examine the effect of the stochastic tail. 
Since the benefit of this component is less significant in simpler environments, we use Humanoid as a representative task for analysis. 
Figure~\ref{fig:ablation_wotail} shows that removing the stochastic tail results in slower convergence than TRFP and also leads to an approximately 10\% lower final return. 
This result suggests that the primary contribution of the stochastic tail lies in improving exploration efficiency in challenging environments.

\section{Conclusion}

In this work, we introduced TRFP, a flow-based MaxEnt RL framework that unifies expressive policy modeling, stable training, and effective one-step sampling by combining a deterministic-stochastic hybrid architecture with surrogate entropy optimization and flow straightening regularization.
Empirical results show that TRFP delivers superior performance across the MuJoCo benchmarks.
More importantly, without an additional distillation procedure, it retains high performance fidelity under one-step sampling.

Several limitations nevertheless remain. In particular, the surrogate log-likelihood used in TRFP is still only an approximation to the true action likelihood.
Moreover, the present study mainly focuses on MuJoCo benchmarks, and thus does not yet validate the framework on larger-scale robot manipulation or long-horizon planning tasks. Future work will explore extending TRFP to broader scenarios, including offline RL and fine-tuning of VLA policies.


\appendix
\section{Hyperparameter Settings}
\begin{table}[!htbp]
\centering
\caption{Detailed Hyperparameter Settings.}
\label{tab:hyperparams}
\setlength{\tabcolsep}{6pt}
\small
\begin{tabular}{ll}
\toprule
Hyperparameter & Value \\
\midrule
Number of denoising steps $K$ & 4 \\
Tail length $L$ & 1 \\
One-step evaluation tail length & 0 \\
Discount factor $\gamma$ & 0.99 \\
Replay buffer size & 1e6 \\
Batch size & 256 \\
Actor learning rate & 3e-4 \\
Critic learning rate & 3e-4 \\
Temperature learning rate & 3e-4 \\
Hidden layers in actor network & [256,256,256] \\
Hidden layers in critic network & [256,256,256] \\
Activation in actor network & Mish \\
Activation in critic network & Mish \\
Straightening regularization weight $\lambda_{\mathrm{fm}}$ & 0.1 \\
Target entropy & -dim($\mathcal{A}$) \\
\bottomrule
\end{tabular}
\end{table}

\bibliographystyle{unsrt}  
\bibliography{ref}  


\end{document}